\documentclass[lettersize,journal]{IEEEtran}
\usepackage{amsmath,amsfonts}
\usepackage{algorithmic}
\usepackage{array}
\usepackage[caption=false,font=normalsize,labelfont=sf,textfont=sf]{subfig}
\usepackage{xcolor}
\usepackage{textcomp}
\usepackage{stfloats}
\usepackage{url}
\usepackage{xcolor}
\usepackage[normalem]{ulem}

\usepackage{verbatim}
\usepackage{graphicx}
\def\BibTeX{{\rm B\kern-.05em{\sc i\kern-.025em b}\kern-.08em
	T\kern-.1667em\lower.7ex\hbox{E}\kern-.125emX}}
\usepackage{balance}
\begin{document}
\title{Granular-Ball Computing: An Efficient, Robust, and Interpretable Adaptive Multi-granularity Representation and Computation Method}
\author{Shuyin Xia, Guoyin Wang, Xinbo Gao, Xiaoyu Lian, Hongzhi Kuai
	\thanks{S. Xia, X. Gao, X. Lian, H. Kuai are with the Chongqing Key Laboratory of Computational Intelligence, 
		Key Laboratory of Cyberspace Big Data Intelligent Security, Ministry of Education, 
		and Key Laboratory of Big Data Intelligent Computing, Chongqing University of Posts and Telecommunications, 400065, Chongqing, China. E-mail: xiasy@cqupt.edu.cn, gaoxb@cqupt.edu.cn, lianxiaoyu724@qq.com, hz\_kuai@163.com}
	\thanks{G. Wang are with the College of Computer and Information Science, the National Center for Applied Mathematics in Chongqing, Chongqing Normal University, Chongqing 401331, China. E-mail: wanggy@cqnu.edu.cn.}
}

\markboth{Journal of \LaTeX\ Class Files,~Vol.~18, No.~9, September~2020}%
{How to Use the IEEEtran \LaTeX \ Templates}

 \maketitle

\begin{abstract}
Human cognition follows a global-first mechanism, enabling adaptive multigranular data representation and processing with inherent efficiency, robustness, and interpretability. Although many existing artificial intelligence methods exhibit certain multigranularity characteristics, their computation still relies primarily on the finest-grained inputs, such as sample points, pixels, and tokens, and thus often suffers from overly fine computational granularity, single-scale modeling, and limited adaptability. To address these limitations, Wang Guoyin and Xia Shuyin et al. proposed granular-ball computing as a new artificial intelligence learning paradigm. Unlike traditional clustering, which mainly serves as a macro-level grouping method, granular-ball computing uses hyperspheres of different sizes, namely granular balls, as meso-level representation units; in low-dimensional spaces, approximate spherical forms such as rectangles and ellipsoids may also be used. With such representations, granular-ball computing can adaptively fit arbitrary data distributions and replace traditional artificial intelligence computation based on fine-grained point inputs or single-granularity modeling, thereby establishing a new theoretical paradigm for AI computation based on granular balls. Its goal is to build an end-to-end multigranular artificial intelligence framework that improves the efficiency, robustness, and interpretability of existing methods. In recent years, this theory has developed rapidly and produced a number of representative results, yet it still lacks a unified model for systematic summarization. To address this gap, this article systematically reviews the major advances in granular-ball computing and, for the first time, proposes a general representation model of granular-ball computing under a unified descriptive framework. On this basis, the article systematically summarizes the fundamental ideas, major research progress, and key methods of granular-ball computing across granular-ball supervised learning, granular-ball unsupervised learning, approximate granular-ball representation and computation, granular-ball deep learning based on latent-space granulation, granular-ball graph learning, and cross-disciplinary theoretical and methodological extensions of granular-ball computing, and further analyzes its current challenges and future development trends.
\end{abstract}


\section{Introduction}

Multi-granularity is a common mode of human cognition and an important organizing principle of the objective world. Intuitively, it is reflected in coarse-to-fine hierarchical structures and their relationships, such as the organization from country to individual or the folder hierarchy in computer systems. In artificial intelligence, many methods also exhibit certain multi-granularity characteristics. For example, in k-means clustering, a larger $k$ corresponds to coarser granularity, whereas a smaller $k$ corresponds to finer granularity; in convolutional neural networks, deeper features are usually more abstract and thus correspond to coarser granularity. Multi-granularity is not only an important way of organizing data, but also provides a natural mechanism for improving model interpretability.

Granular computing is an important theoretical direction in artificial intelligence with a long research history. Zadeh introduced the concept of granular computing in 1979~\cite{zadeh1979fuzzy,zadeh1997toward}. Its early core theories mainly include fuzzy sets~\cite{backer1981clustering,kosko1986counting}, rough sets~\cite{hu2017large,liang2012group,qian2010positive}, quotient space theory~\cite{ling2003theory,zhang2004quotient}, and cloud models~\cite{li1998uncertainty,li2009new}. Among them, fuzzy sets characterize fuzzy granulation through membership functions, whereas rough sets and quotient space theory construct information granules at different scales through equivalence relations. Quotient space theory and cloud models are also important original contributions from Chinese researchers in information science. Around rough-set-related directions, Chinese scholars have further made substantial contributions, including studies on the differences between algebraic and information-theoretic formulations of rough sets~\cite{wang2003rough}, the equivalence between fuzzy soft sets and fuzzy information systems~\cite{pei2005soft}, positive-region-based reduction and core attribute acceleration~\cite{qian2010positive}, topological properties of covering rough sets~\cite{xu2005properties}, the influence of label noise on lower and upper approximations~\cite{yao2007decision}, neighborhood and robust rough sets~\cite{hu2008neighborhood,hu2011robust}, multi-scale decision tables~\cite{wu2011theory}, variable-precision rough-set clustering~\cite{han2006novel}, and incremental approximation in dynamic environments~\cite{chen2011rough}. Building on these developments, Xia et al. further studied local attribute redundancy, robustness of rough sets, rough-set-based classification, and concept-lattice organization~\cite{xia2022efficientrough}, and proposed granular-ball rough sets~\cite{xia2022grrs,xia2023gbrs}, which enable equivalence-class-based analysis of continuous data and provide a unified description of Pawlak rough sets and neighborhood rough sets.

Early studies by Chen et al. showed that human cognition follows a ``global-first'' principle, in which the visual system first captures the overall topological structure and then gradually refines local details~\cite{chen1982topological}. Inspired by this cognitive mechanism, Wang proposed data-driven granular cognitive computing~\cite{wang2017dgcc}, aiming to develop an artificial intelligence paradigm more consistent with human cognition. Xia and Wang et al. further argued that such a coarse-to-fine cognitive process is inherently efficient, robust, and interpretable, whereas many existing AI methods still rely primarily on fine-grained inputs and fixed-scale relationships, leading to a clear mismatch with human cognition. Although methods such as classification, clustering, nearest-neighbor analysis, rough sets, and deep learning may exhibit certain multi-granularity characteristics in their outcomes, their computational processes are still largely based on the finest-grained inputs, such as sample points, pixels, or tokens, and usually operate at fixed scales. As a result, they remain limited in efficiency, robustness, adaptability, and interpretability. Therefore, developing a truly multi-granular, efficient, and adaptive representation and computation model has become an important scientific problem for advancing artificial intelligence, especially in the era of foundation models.

Although prior studies have introduced multi-granularity ideas into classification learning~\cite{al2014data,roh2014design}, most of them still regard granulation as a preprocessing step rather than a unified computational paradigm built on multi-granularity input units. Ideally, a multi-granularity representation mechanism should be efficient, adaptive, and weakly dependent on downstream learning; otherwise, its generation process may itself become a bottleneck. Existing methods, including those based on k-means, density clustering, spectral clustering, manifold clustering, clustering SVMs, data-granulation SVMs, and granular neural networks~\cite{barros2000SVM-KM,pedrycz2001granular,syeda2002parallel,tang2012granular}, still essentially operate within a point-wise analysis paradigm centered on individual samples.
 
Against this background, to enable a more thorough end-to-end multi-granularity computation paradigm while simultaneously improving the efficiency, robustness, and interpretability of artificial intelligence methods, Xia and Wang et al.~\cite{xia2019granular} proposed a new AI computing paradigm, namely granular-ball computing. Its core idea is to replace the finest-grained sample points as the basic input and computational units of learning models with coarse-grained, adaptive, and multi-granular information granules, thereby establishing a genuinely unified learning paradigm based on multi-granularity inputs. Spheres are adopted as the basic information granules mainly because they admit simple, unified, and symmetric mathematical representations in arbitrary-dimensional spaces. A standard granular ball can be represented as
\begin{equation}
	\{x \in \mathbb{R}^d \mid |x-c| \le r\},
\end{equation}
where $x$ denotes a sample within the granular ball, and $c$ and $r$ denote its center and radius, respectively. $d$ is the dimensionality. Since a granular ball can be represented by only these two parameters regardless of dimensionality, it is particularly suitable for high-dimensional data modeling; moreover, its continuously differentiable form also provides a basis for optimization and learning in related models. In low-dimensional spaces, ellipsoids, rectangles, and other approximate shapes may also be adopted to better fit local data distributions.

From the perspective of representation hierarchy, a granular ball is a meso-level representation between microscopic sample points and macroscopic global structures. By using regional units with a certain coverage range and local consistency, it bridges local structure characterization and global structure understanding. On the one hand, coarse-grained granular balls replace traditional point-wise or pixel-wise inputs; since the number of granular balls is usually far smaller than the number of sample points, granular-ball computing is more efficient. At the same time, by aggregating local samples, granular balls can effectively reduce the influence of individual noisy points and outliers, thereby improving robustness. On the other hand, the granularity of granular balls can be adaptively determined according to the data distribution, leading to a more complete multigranular computing paradigm and enabling the construction of local equivalence classes and hierarchical structures with topological meaning in continuous spaces, which further enhances interpretability.

\begin{figure*}[ht!]
	\centering
	\includegraphics[width=6.5in]{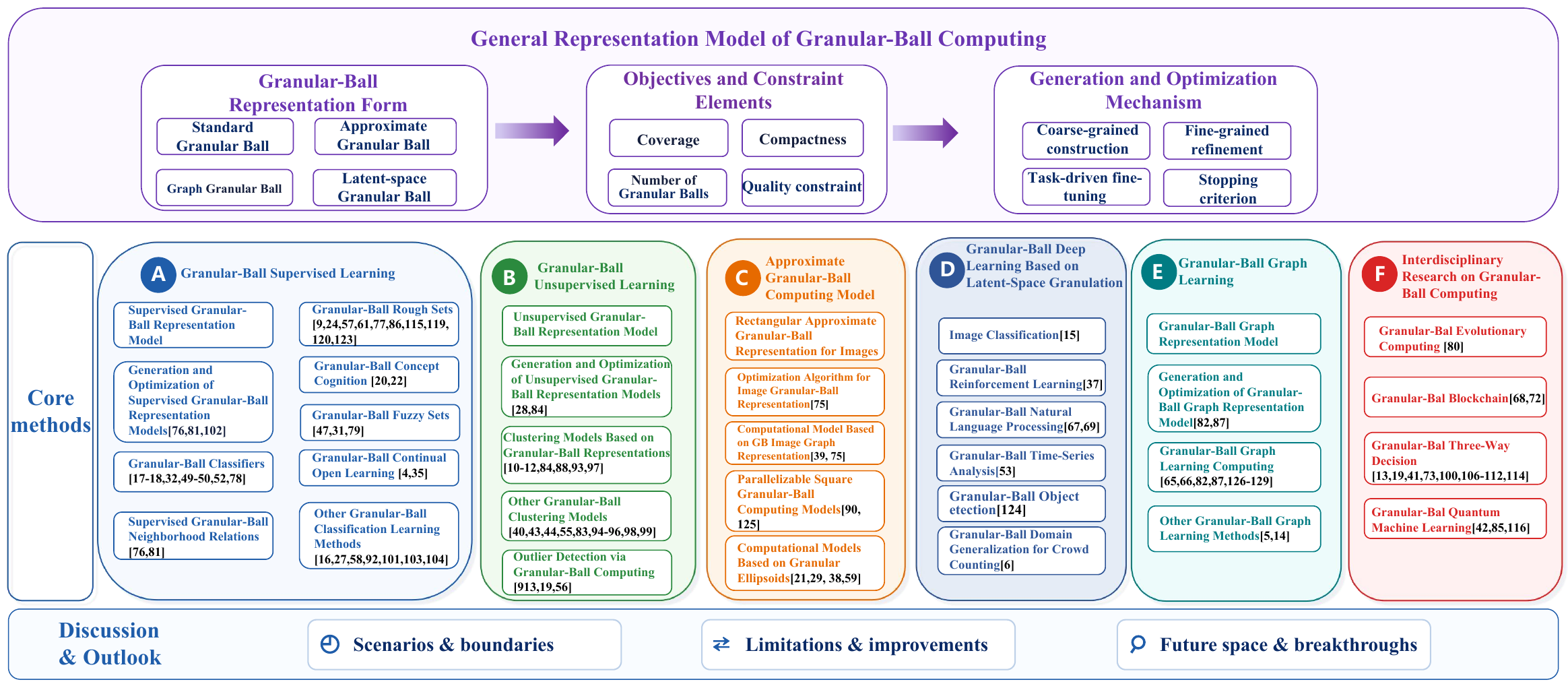}
	\caption{Review Framework for Granular-Ball Computing.}
	\label{FigGB}
\end{figure*}

In recent years, granular-ball computing has been progressively extended to multiple directions, including classification~\cite{quadir2024granular,xia2024gbsvm,xia2024fuzzy,xia2019granular}, clustering~\cite{GBCT2024,xie2025tpami}, rough sets~\cite{xia2023gbrs}, graph learning~\cite{SGBGC2025,GBGC2025}, deep learning~\cite{xia2025graphRespresation}, evolutionary optimization~\cite{xia2023granular}, and quantum machine learning~\cite{QGBKNN2025}, leading to a series of representative advances. However, existing studies remain scattered, and there is still a lack of a survey that systematically summarizes the fundamental ideas, core mechanisms, and methodological framework of granular-ball computing. Given that granular-ball methods across different directions share clear commonalities in representation form, generation mechanism, and computation pattern, this article, for the first time, organizes the related studies under a unified descriptive modeling framework into granular-ball representation models and their optimization, granular-ball supervised learning, granular-ball unsupervised learning, approximate granular-ball representation and computation, granular-ball deep learning based on latent-space granulation, granular-ball graph learning, and cross-disciplinary theoretical and methodological extensions of granular-ball computing, as illustrated in Fig.~\ref{FigGB}.

The main contributions of this survey are as follows.

\begin{itemize}
	\item{It systematically clarifies the motivation, background, and core role of granular-ball computing, showing how it has evolved from the global-first mechanism of human cognition into a multigranular learning paradigm that uses granular balls as meso-level representation units for efficient, robust, and interpretable intelligent computing.}

    \item{For the first time, this work establishes a general representation model of granular-ball computing. The model provides a unified characterization of granular-ball methods in terms of representation form, quality constraints, generation and optimization, and computation pattern, thereby offering a theoretical framework for the unified analysis and comparison of granular-ball methods across different directions.}

   \item{Based on the above unified model, it categorizes existing studies into granular-ball supervised learning, granular-ball unsupervised learning, approximate granular-ball representation and computation, granular-ball deep learning based on latent-space granulation, granular-ball graph learning, and cross-disciplinary theoretical and methodological extensions of granular-ball computing, and systematically summarizes the representative advances, key ideas, and intrinsic connections among these directions.}

   \item{It further analyzes the major challenges and applicability boundaries of granular-ball computing, and discusses its potential future directions in large models, trustworthy artificial intelligence, and complex open environments, thereby providing references for subsequent theoretical research, method design, and application development.}
	
\end{itemize}

The remainder of this article is organized as follows. Section 2 introduces the basic concepts, representation forms, and generation mechanisms of granular-ball computing. Section 3 reviews its advances in supervised learning, while Section 4 summarizes the main methods in unsupervised learning. Section 5 presents approximate granular-ball representations and computational models. Section 7 discusses granular-ball representation and learning for graph-structured data. Section 8 summarizes cross-disciplinary theoretical and methodological extensions of granular-ball computing. Section 9 further discusses its major applicable scenarios, current limitations, and future research directions.

\section{Granular-Ball Representation Models and Optimization}

\subsection{Basic Model of Granular-Ball Representation}

The core issue of granular-ball representation is how to characterize a data representation composed of multiple granular balls in a unified and computable form. From a modeling perspective, granular-ball representation is not merely a simple aggregation of samples, but must jointly balance coverage, compactness, the number of granular balls, and granular-ball quality. The granular-ball set should cover the original samples as much as possible to reduce information loss; each granular ball should remain relatively compact to ensure that a coarse-grained unit still carries clear semantics and structure; and the number of granular balls determines the granularity of the representation, with too many leading to point-level degeneration and too few resulting in over-coarsening and an inability to capture the true distribution. Therefore, the generation and optimization of granular-ball representation are essentially a trade-off among coverage, compactness, the number of granular balls, and granular-ball quality.

Given a dataset $X=\{x_1,x_2,\ldots,x_n\}$, let $GB=\{GB_1,GB_2,\ldots,GB_m\}$ denote the set of granular balls generated on $X$. In general, a granular ball is represented in the standard spherical form as $GB_i=(X_i,c_i,r_i)$, where $X_i$ is the sample set contained in $GB_i$, and $c_i$ and $r_i$ denote its center and radius, respectively~\cite{xia2019granular}. The standard spherical form is the only granular-ball representation with good scalability in high-dimensional spaces. In low-dimensional spaces, however, non-standard representations such as rectangles~\cite{xia2025graphRespresation} and ellipsoids~\cite{kerbl20233d} can also be adopted. The basic representation model of granular-ball computing is defined as follows:
\begin{equation}\label{equ:GBC_1}
	\begin{aligned}
		& \min_{c_i,r_{i},m} \  -Cov(GB)-Comp(GB) + m \\
		& s.t.  \ \ quality(GB_i) \geq \phi(X),i = 1,2,...,m,     \\
	\end{aligned}
\end{equation}
where $Cov(GB)$ denotes the coverage of the granular ball set over the samples. $Comp(GB)$ denotes the compactness of granular balls and $m$ is the number of granular balls. $\phi(X)$ is a decision function for granular-ball quality constraints, which specifies a lower bound on granular-ball quality. It can be defined either as a constant threshold or as an adaptive function. Common ways of setting this threshold include preset values, grid search, and adaptive adjustment, among which adaptive schemes are generally more consistent with the goal of reducing manual intervention in granular-ball computing~\cite{xia2022efficient}.

Assume that granular ball $GB_i$ contains $N$ samples. Its center is usually defined as the mean of the samples within the ball:
\begin{equation}\label{center}
	c_i=\frac{1}{N}\sum_{j=1}^{N}x_j.
\end{equation}
Two common definitions of the granular-ball radius are
\begin{equation}\label{r}
	r^{ave}_i=\frac{1}{N}\sum_{j=1}^{N}{\|x_{j} - c_i\|},r^{max}_i={\max_{{x_{j}}\in GB_i}\left \{ {\| x_{j} - c_i\|}\right \}},
\end{equation}
where $\|x_{j} ,c_i\|$ denotes the distance between sample $x_{j}$ and the center $c_i$. The former is the average distance, which yields a clearer local boundary and is commonly used in supervised learning; the latter uses the maximum distance and emphasizes the coverage ability of a granular ball over the samples.

\begin{figure}[ht!]
	\centering
	\includegraphics[width=0.9in]{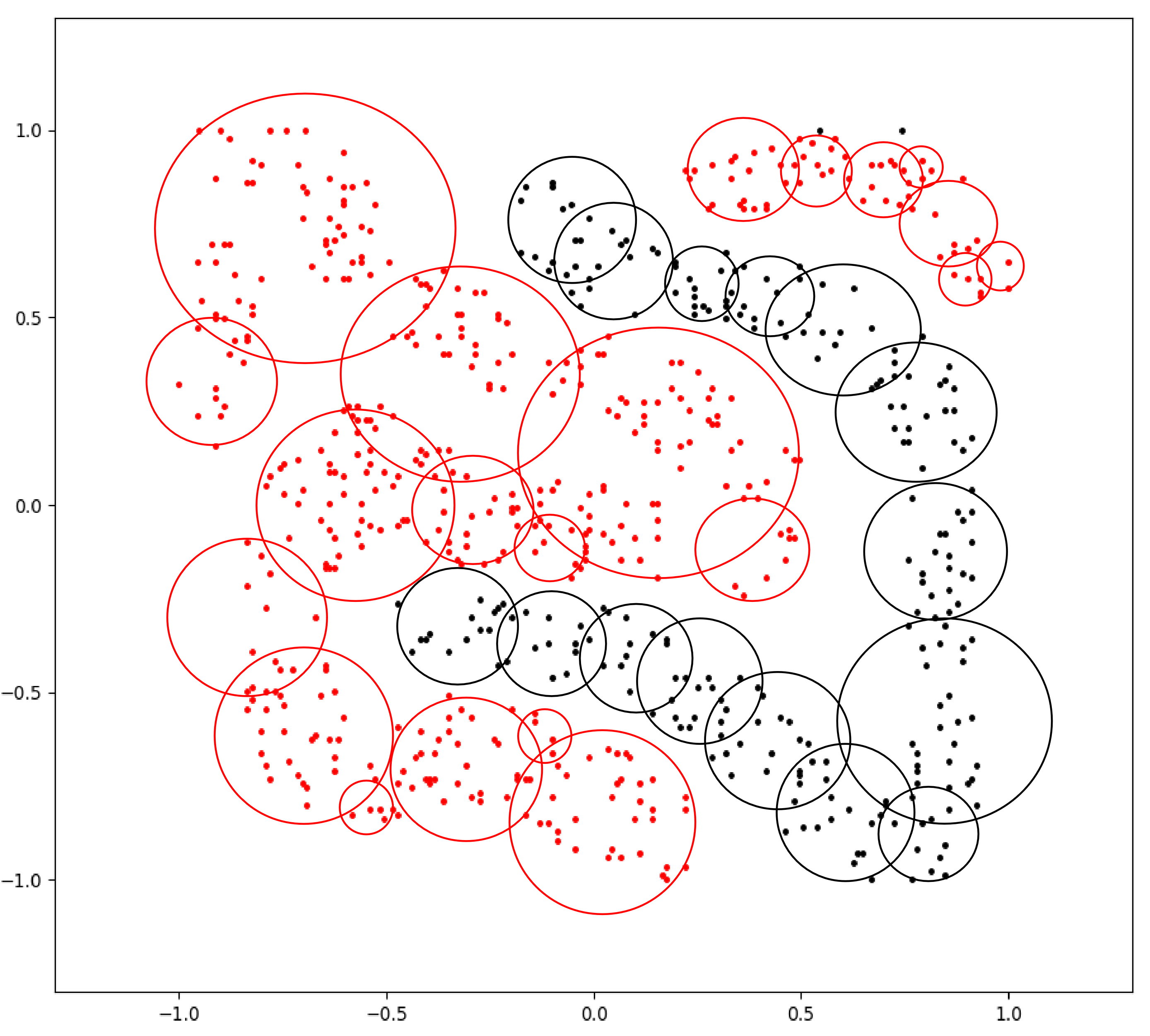}
	\includegraphics[width=0.9in]{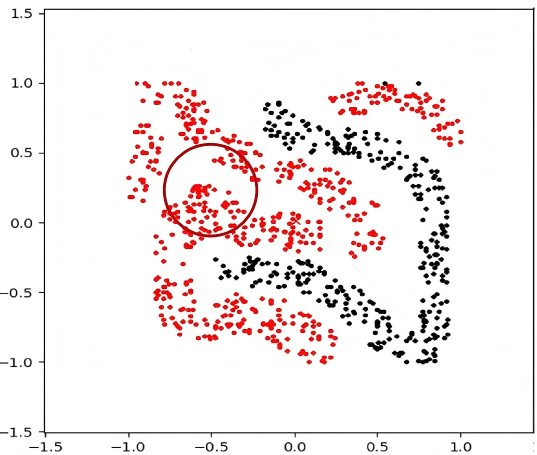}
	\includegraphics[width=0.9in]{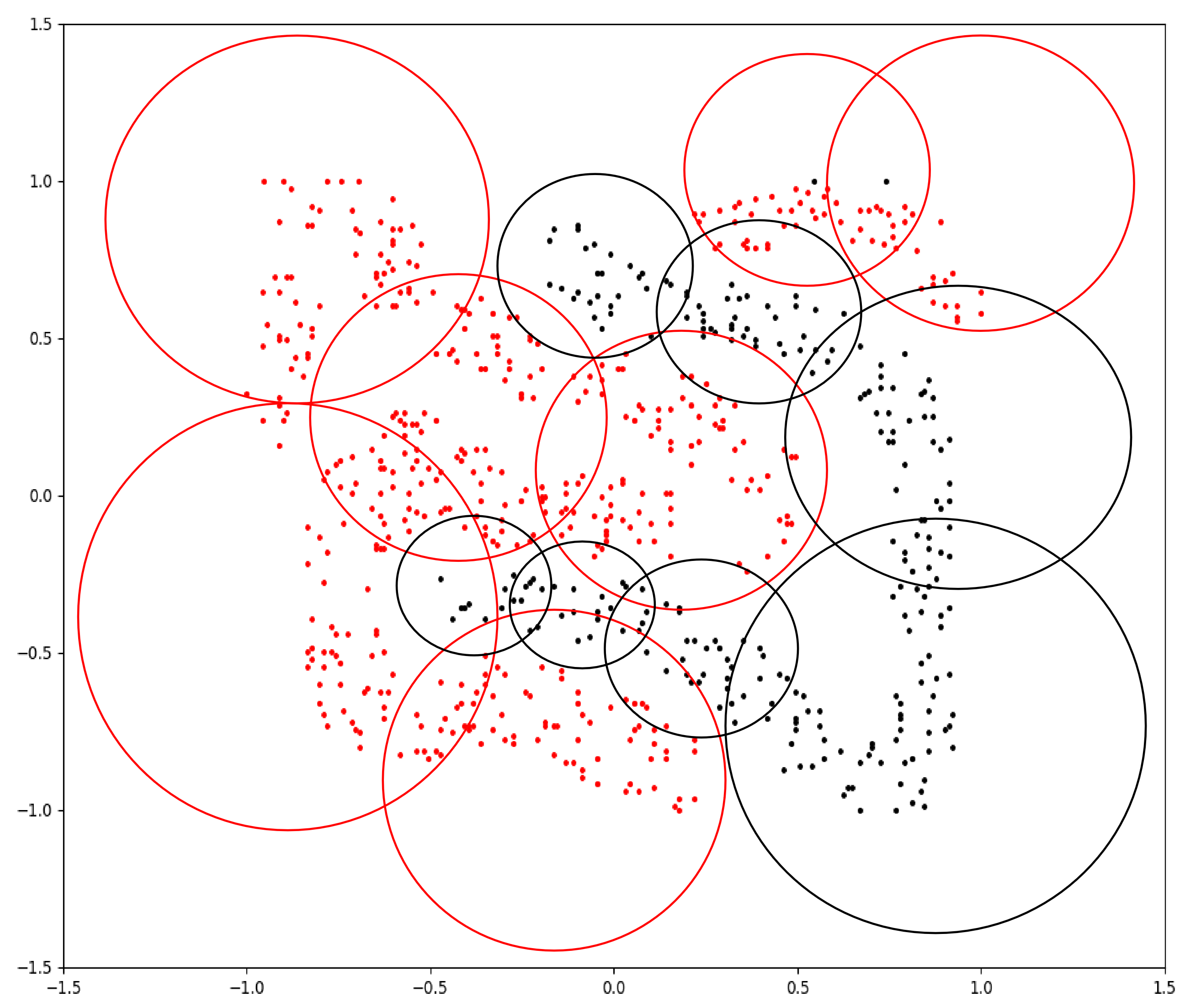}\\
	\quad(a)\quad\quad\quad(b)\quad\quad\quad\quad\quad(c) \\
	\includegraphics[width=1.2in]{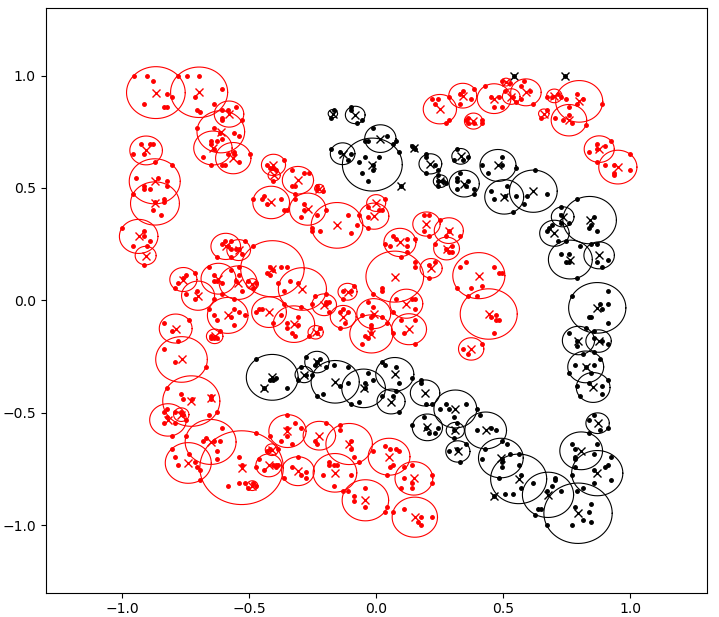}
	\includegraphics[width=1.2in]{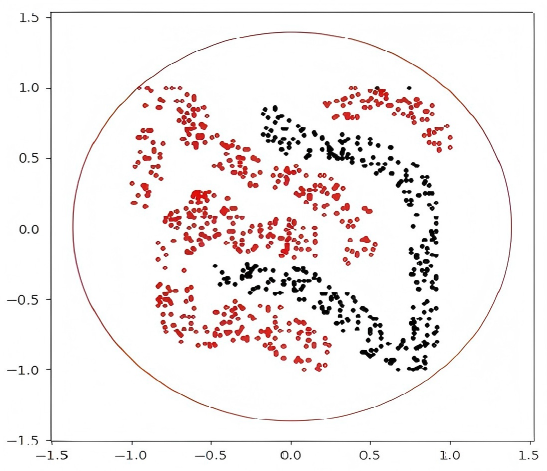}\\
	(d)\quad\quad\quad\quad\quad\quad\quad\quad(e) \\
	\caption{ Analysis of the granular-ball generation model using classification as an example.(a)The granular-ball coverage result when all elements in the model work together;(b)without the coverage term; (c)without the compactness term;(d)without the term for the number of granular balls;(e) without the quality constraint. }
\label{fig:GBmodel}
\end{figure}

Taking traditional classification models, such as granular-ball support vector machines, as an example, these methods usually do not involve transformations of the feature representation space. In the objective function of the representation model in Eq.~(\ref{equ:GBC_1}), the coverage is typically set to 1, meaning that the granular-ball set is required to cover all samples. The radius is defined by the farthest distance to ensure compactness, thereby satisfying a fixed compactness term. In this case, the granular-ball representation model in Eq.~(\ref{equ:GBC_1}) can be further simplified as follows:
 \begin{equation}\label{equ:Supervised}
 	\begin{aligned}
 		& \min_{c_i,r_i,m} \ m \\
 		\text{s.t.} \quad & quality(GB_i) \ge T,\ i=1,2,\ldots,m, \\
 		& dis(c_i,c_j) \ge r_i+r_j,\ \text{if}\ l(GB_i)\ne l(GB_j), \\
 		& c_i=\frac{1}{N}\sum_{j=1}^{N}x_j,\quad r_i=\max_{x_{ij}\in GB_i}{ \| x_{ij}- c_i \|}.
 	\end{aligned}
 \end{equation}
Here, $l(GB_i)$ denotes the label of granular ball $GB_i$. Based on the basic representation model in Eq.~(\ref{equ:GBC_1}), achieving a desirable granular-ball representation as shown in Fig.~\ref{fig:GBmodel}(a) requires all components to work together; the absence of any component will lead to imbalance in the system: 
 
 (1) \textbf{Coverage term}: With other factors fixed, higher coverage implies less information loss and a more complete characterization of the data distribution. Without the coverage term $Cov(GB)$, some samples may not be effectively represented, as shown in Fig.~\ref{fig:GBmodel}(b).
 
 (2) \textbf{Compactness term}: Compactness requires the samples within each granular ball to be as concentrated as possible. Without the compactness term $Comp(GB)$, granular balls may become overly large and internally loose, making it difficult to characterize the data distribution accurately and weakening interpretability, as shown in Fig.~\ref{fig:GBmodel}(c).
 
 (3) \textbf{Number of granular balls}: The number of granular balls determines the granularity of the representation. Minimizing $m$ helps maintain a coarse-grained representation and improves efficiency and robustness. Without the term $m$, granular balls may become excessively many and overly small, as shown in Fig.~\ref{fig:GBmodel}(d), and in the extreme case may even degenerate into single-sample balls.
 
 (4) \textbf{Quality constraint}: $\phi(X)$ may be specified as a fixed threshold or an adaptive function to meet different task requirements. Without such a quality constraint, a degenerate solution may arise in which a single granular ball covers all samples; in this case, the resulting representation is of poor quality and cannot capture any meaningful distributional characteristics of the data, such as decision boundaries in classification, as shown in Fig.~\ref{fig:GBmodel}(e). This constraint can also be combined with theoretical frameworks such as the ``justifiable granularity'' principle to guide the quality function toward a balance between coverage and specificity~\cite{GBJRC2025,yang2023granular}.

The four components in the granular-ball representation model given in Eq.~(\ref{equ:GBC_1}) jointly determine the rationality of the resulting representation. Specifically, coverage and compactness emphasize adequate representation of the samples and concentration within each granular ball, respectively, while the number of granular balls and the quality constraint together balance representation granularity and quality. These four factors therefore constitute the foundation of effective multi-granular representation in granular-ball computing. In many specific tasks, this model can be further simplified. For example, it can be simplified into the granular-ball classification model shown in Eq.~(\ref{equ:Supervised}).


%

\subsection{Generation and Optimization of the Basic Granular-Ball Representation Model}

The granular-ball representation model (\ref{equ:Supervised}) requires the joint determination of the number, centers, and radii of granular balls, making direct optimization generally nonconvex and computationally challenging. To address this, granular-ball computing adopts a coarse-to-fine generation strategy consistent with the global-first principle of human cognition, which efficiently approximates the optimal representation under quality constraints and adaptively determines the size, number, and granularity of granular balls through progressive splitting. The time complexity of this process is typically below $O(n^2)$, and in some cases can even be linear or sublinear, thereby providing the basis for the advantages of granular-ball computing in efficiency, robustness, interpretability, and generalization. Taking classification as an example, a common granular-ball generation method~\cite{xia2019granular} sets the quality threshold $T$ in Eq.~(\ref{equ:Supervised}) to 1, starts from an initial granular ball covering all samples, and then progressively refines it through splitting functions such as partitioning or $k$-means until all granular balls satisfy the quality threshold. It should be noted that this is only the basic generation framework of granular-ball computing; the specific quality measure, splitting strategy, and stopping criterion can all be flexibly replaced or extended according to task requirements~\cite{xie2023gbg}.

Although Eq.~(\ref{equ:GBC_1}) provides a reasonable description of granular-ball representation in relatively simple settings such as classification and clustering, real-world data often exhibit more complex characteristics, including multimodality, noise contamination, and distributional imbalance, under which its fixed objective and constraint form may become insufficient. This calls for a more general granular-ball representation model that can adapt its representation form, quality criterion, and generation mechanism according to data characteristics and task requirements, thereby supporting stable, efficient, and interpretable multi-granular learning in more complex scenarios.

\subsection{General Model of Granular-Ball Representation}

To describe a broader granular-ball learning framework, the generalized representation model of granular-ball computing needs to be extended in several aspects. First, in addition to the standard hypersphere commonly used in high-dimensional spaces, non-standard geometric structures such as rectangles~\cite{xia2025graphRespresation} and ellipsoids~\cite{guo2026granular,kerbl20233d,sun2025gec} can also be adopted in low-dimensional settings according to task characteristics. This motivates the introduction of an attribute parameter vector $\vec{\theta_i}$ to uniformly represent the radius, side lengths, axis lengths, and other statistical parameters, yielding $GB_i=(\varphi(X_i),c_i,\vec{\theta_i})$. Second, coverage and compactness can be described more flexibly by incorporating principles such as justifiable granularity, and thus a function $\mathcal{J}(-Cov(GB),-Comp(GB))$ is introduced to support more general granular-ball evaluation objectives. In addition, in settings such as deep learning, granular-ball structures often need to be optimized jointly with model parameters. The generalized model therefore includes $\vec{\beta}$ as optimization variables and uses $\varphi(X)$ to denote samples in the latent space. Finally, depending on the task, granular-ball generation may also be subject to additional constraints. For example, in classification, heterogeneous granular balls may be required not to overlap, which can be uniformly expressed as $C(GB,X)\geq 0$. Based on these extensions, the generalized representation model of granular-ball computing can be formulated as follows:
\begin{equation}\label{equ:GBC}
	\begin{aligned}
		&\quad \quad \quad \quad f(\varphi(X),\vec{\alpha}) \longrightarrow g(GB,\vec{\beta})\\
		&\min_{c_i,\vec{\theta_i},\vec{\beta},m} \ \mathcal{J}(-Cov(GB),-Comp(GB)) + m \\
		& s.t. \ \ quality(GB_i) \geq \phi(\varphi(X)), \ i=1,2,\ldots,m \\
		&\ \ \ \ \ \ C(GB,\varphi(X))\geq 0,
	\end{aligned}
\end{equation}
where $f(\varphi(X),\vec{\alpha})$ denotes a computational model with input $\varphi(X)$ and parameters $\vec{\alpha}$; when granularization is performed on the original samples, $\varphi(X)=X$. $g(GB,\vec{\beta})$ denotes a granular-ball computing model that takes $GB$ as input and is parameterized by $\vec{\beta}$. Compared with the basic representation model Eq.~(\ref{equ:GBC_1}), the optimization variables in Eq.~(\ref{equ:GBC}) are extended from the number, centers, and radii of granular balls to the number of granular balls, centers, attribute parameters $\vec{\theta_i}$, and granular-ball model parameters $\vec{\beta}$, which can be optimized separately or jointly depending on the task. The constraint term $C(GB,\varphi(X))$ is used to characterize relations among granular balls and samples, such as overlap, inclusion, and exclusion, thereby enhancing the expressive power and adaptability of the model.

Eq.~(\ref{equ:GBC}) reflects a new artificial intelligence paradigm, in which traditional point-wise learning is replaced by a multi-granular computing framework based on granular balls. Its key idea is that the basic computational units are no longer the finest-grained sample points, which often lack semantic meaning, but granular balls endowed with multi-granularity and equivalence-class characteristics. By leveraging the adaptive multi-granular representation capability of granular balls, this model provides more semantic and interpretable data representations and further gives rise to new computational models and methods.

\subsection{Generation and Optimization of the General Granular-Ball Representation Model}

\begin{figure}[ht!]
	\centering
	\includegraphics[width=3.5in]{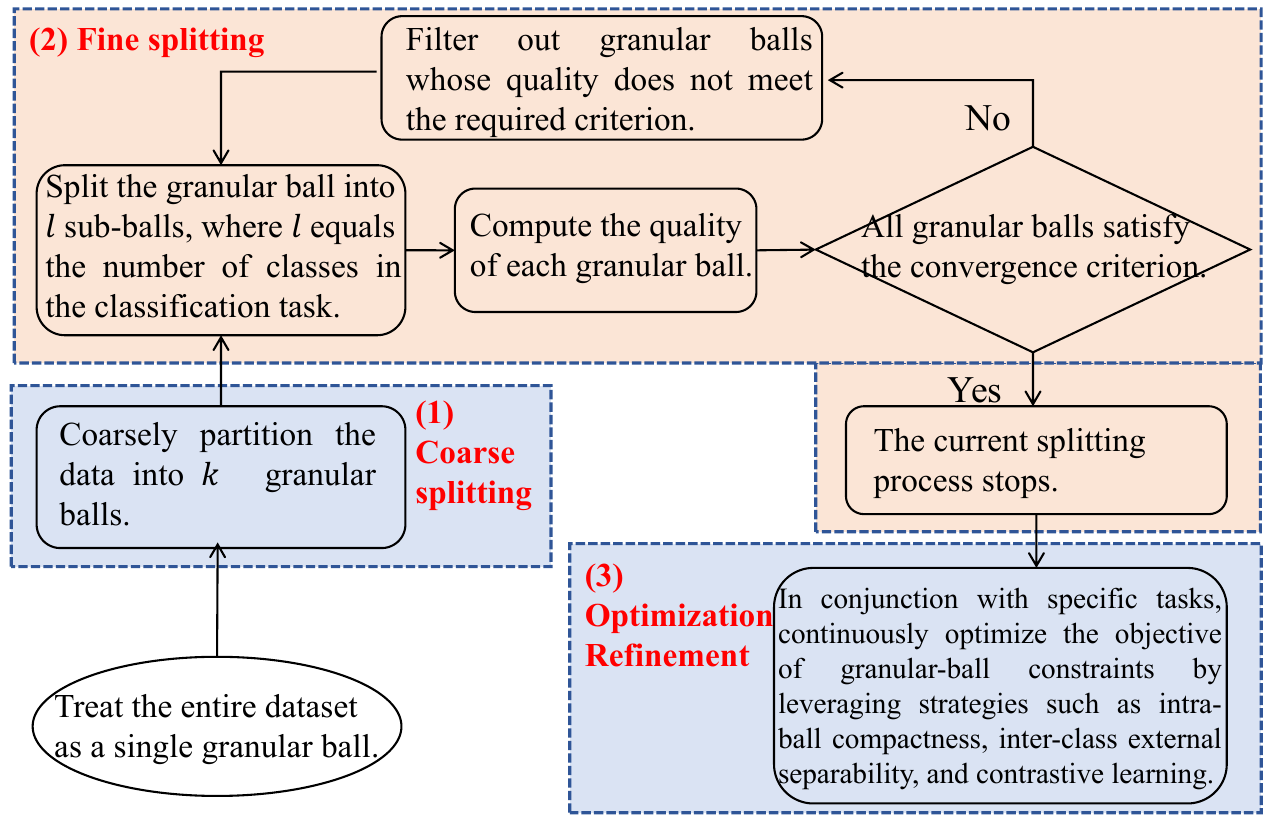}
	\caption{ Optimization process of the general granular-ball generation model.}
	\label{fig:GBgeneration_2}
\end{figure}

In settings such as deep learning, granular-ball structures evolve dynamically with model parameters and the training process, and their generation and optimization extend the basic model by introducing an additional adjustment stage. In particular, an early coarse splitting stage is often recommended to obtain a better global initialization and simplify complex data distributions, thereby improving the effectiveness of subsequent quality evaluation and reducing the risk of premature convergence in adaptive splitting. Excessive coarse splitting may increase computational cost, whereas insufficient splitting may fail to achieve the desired simplification; therefore, $\sqrt{n}$ is often adopted as a practical compromise, although this step is not mandatory. Specifically, granular-ball generation can be divided into three stages, as illustrated in Fig.~\ref{fig:GBgeneration_2}: (1) Coarse-grained construction, which provides an initial macroscopic partition of the data; (2) Fine-grained refinement, in which granular balls that do not satisfy the quality criterion are recursively split to improve local representation accuracy and compactness; and (3) Task-driven fine-tuning, in which the coverage, compactness, and number of granular balls are jointly optimized under specific learning objectives, especially in deep learning scenarios. This coarse-to-fine process is consistent with the global-first principle of human cognition: first capturing the global structure, then refining local details, and finally adjusting the representation according to task requirements. It should be emphasized that granular-ball optimization is essentially a general framework for describing the core process of granular-ball generation and iterative refinement, rather than a rigid specification of quality metrics, split/merge strategies, or stopping criteria. In practice, it can be flexibly extended according to data characteristics and task requirements, for example by designing more suitable quality functions, adopting different split and merge rules, or integrating it with other learning mechanisms to further improve efficiency, robustness, and adaptability.

\textbf{Discussion: Differences and Connections Between Granular-Ball Representation/Generation and Traditional Clustering}

Granular balls provide a meso-level data representation that differs fundamentally from traditional clustering, which can largely be viewed as a macro-level representation scheme. First, granular-ball computing can adaptively form multi-granular structures to fit complex data distributions: smaller granular balls are generated in structurally complex regions, whereas larger ones are used in simpler regions. In contrast, traditional clustering methods, as macro-level models, are generally unable to achieve such flexible fitting of arbitrary data distributions. Second, the radii, sample sizes, and number of granular balls are typically determined adaptively from the data, often requiring only a few or even no additional granularity parameters~\cite{GBCT2024,xia2022efficient}. In addition, granular-ball generation avoids extensive pairwise point-distance computation and usually has sub-quadratic complexity. The standard granular-ball form is also continuous and differentiable, making it amenable to optimization and gradient-based derivation, a property that traditional clustering methods typically do not possess. Finally, granular-ball generation is not restricted to unsupervised settings; in tasks such as classification, label information can be naturally incorporated to produce supervised granular-ball representations. Granular-ball generation is therefore not merely a global grouping process, but an efficient, adaptive, and multi-granular meso-level representation mechanism. Accordingly, granular-ball computing also differs from the conventional paradigm in which clustering serves only as a preprocessing step and subsequent computation remains point-wise. Instead, it directly takes granular balls as the basic input units and establishes a new multi-granular computing paradigm with stronger potential in efficiency, robustness, generalization, and interpretability.

\section{Granular-Ball Supervised Learning}

In supervised learning, classical methods such as classifiers and fuzzy sets often suffer from limited efficiency, robustness, and interpretability due to their reliance on point-wise inputs. Granular-ball computing addresses this issue by adaptively generating multi-granular granular balls according to data distributions and replacing traditional point-level inputs with granular balls, thereby giving rise to a series of supervised granular-ball learning models.

\subsection{Supervised Granular-Ball Representation Model}

In supervised learning tasks, granular balls are typically represented as standard spheres that cover the sample space and serve as the basic input units for subsequent models, enabling multi-granular representation and computation. When supervision is available, the label of a granular ball is usually determined by the majority label of the samples it contains. As a result, a granular ball not only carries statistical information such as center, radius, and label, but also conveys explicit discriminative semantics, making it naturally compatible with supervised models such as classifiers, rough sets, and fuzzy sets. In most supervised settings, granular-ball representation adopts a full-coverage form, while a certain level of compactness is ensured through the radius definition in Eq.~(\ref{r}). Based on the generalized granular-ball representation model (\ref{equ:Supervised}), a commonly used supervised granular-ball representation model can be written as
\begin{equation}\label{equ:Supervised_1}
	\begin{aligned}
		& \min_{c_i,r_i,m} \ m \\
		s.t. \quad & quality(GB_i) \ge T, \ i=1,2,\ldots,m, \\
		& dis(c_i,c_j) \ge r_i+r_j, \ \text{if} \ \ l(GB_i)\ne l(GB_j), \\
		& quality(GB_i)=\frac{\max_j\{l_j\}}{|GB_i|}, \ j=1,2,\ldots,k.
	\end{aligned}
\end{equation}
Here, $l_j$ denotes the number of samples in the granular ball whose label is $j$. The core idea of this model is to cover the sample space with as few granular balls as possible while satisfying quality constraints, where the number of granular balls determines the granularity of the representation, and the quality constraint prevents the representation from degenerating into a small number of low-quality large balls. In classification tasks, granular-ball quality is usually measured by purity: the higher the purity, the more semantically consistent and discriminative the granular ball. When the purity reaches 1, the granular ball becomes a pure ball, which is often regarded as an ideal stopping condition for splitting. The quality constraint $\phi(X)$ can either be specified directly by a constant threshold $T$ or adjusted adaptively according to the local data distribution. Furthermore, according to the generalized representation model (\ref{equ:GBC}), this model also introduces a non-overlap constraint for heterogeneous granular balls, which helps avoid ambiguity at decision boundaries and enhances class separability.

\subsection{Generation and Optimization of Supervised Granular-Ball Representation Models}

Existing granular-ball generation methods can be broadly divided into two categories: sequential generation and splitting-based generation, with the latter being more widely used. Such methods follow the global-first principle of human cognition: they start from the smallest possible number of granular balls and progressively refine them only when the constraints are not satisfied. This process not only reflects a coarse-to-fine cognitive computing paradigm, but also corresponds to the objective of minimizing the number of granular balls in the supervised representation model (\ref{equ:Supervised_1}). In early splitting-based methods, efficient $k$-means was commonly used for splitting, where $k$ is the number of classes within a granular ball~\cite{xia2019granular}, so as to preserve the overall efficiency of granular-ball classifiers. In 2022, Xia et al. further replaced $k$-means with a partition-based strategy, improving the efficiency of granular-ball generation by several to tens of times and reducing the time complexity to $O(mn)$, where $m$ is the number of granular balls and $n$ is the number of samples~\cite{xia2022efficient}. To address the instability caused by random initialization in earlier methods, Xie et al. proposed a stable sequential granular-ball generation strategy in 2023, which directly uses the center of the majority class within a granular ball as the initial center of a child granular ball, thereby improving both stability and convergence speed~\cite{xie2023gbg}. These improvements not only enhance the performance of supervised granular-ball generation, but also provide useful insights for granular-ball construction in other granular-ball computing methods.

\subsection{Granular-Ball Classifiers}

Traditional classifiers usually take fine-grained sample points as input and therefore suffer from limitations in efficiency, robustness, and interpretability. In contrast, granular-ball classifiers use granular balls as the basic input units, transforming classification from point-wise optimization to computation over a set of granular balls. The granular-ball support vector machine (GBSVM) is one of the earliest non-point-input classification models~\cite{xia2024gbsvm}. It replaces sample points with granular balls as the input to SVM, so that the separating hyperplane is determined by support vector balls rather than support vectors. As illustrated in Fig.~3, for any granular ball, the distance from its center to the supporting hyperplane should be greater than its radius, which leads to the primal optimization model of granular-ball SVM:
\begin{equation}\label{equ:oobj-convex}
	\begin{aligned}
		\substack{\min\\\omega,b}\quad\quad &\frac{1}{2} \left\| \omega  \right\|^2\\
		s.t.\quad\quad &y_i(\omega \cdot c_i+b)-\left\| \omega  \right\| r_i\ge 1 \text,\quad i=1,2,\cdots.
	\end{aligned}
\end{equation}
GBSVM is the first multi-granular classifier that does not rely on point-wise inputs and contains no explicit $x_i$. When $r_i \to 0$, $c_i$ can be regarded as an input point, and the model reduces to the traditional SVM. Subsequent studies further developed its dual formulation, optimization procedure, granular-ball generation in kernel space, and granular-ball kernel SVM variants~\cite{xia2024gbsvm}. Experimental results show that GBSVM outperforms traditional SVM in training efficiency, decision efficiency, and robustness to label noise.

\begin{figure}[htbp!]
	\centering
	\includegraphics[width=1.5in]{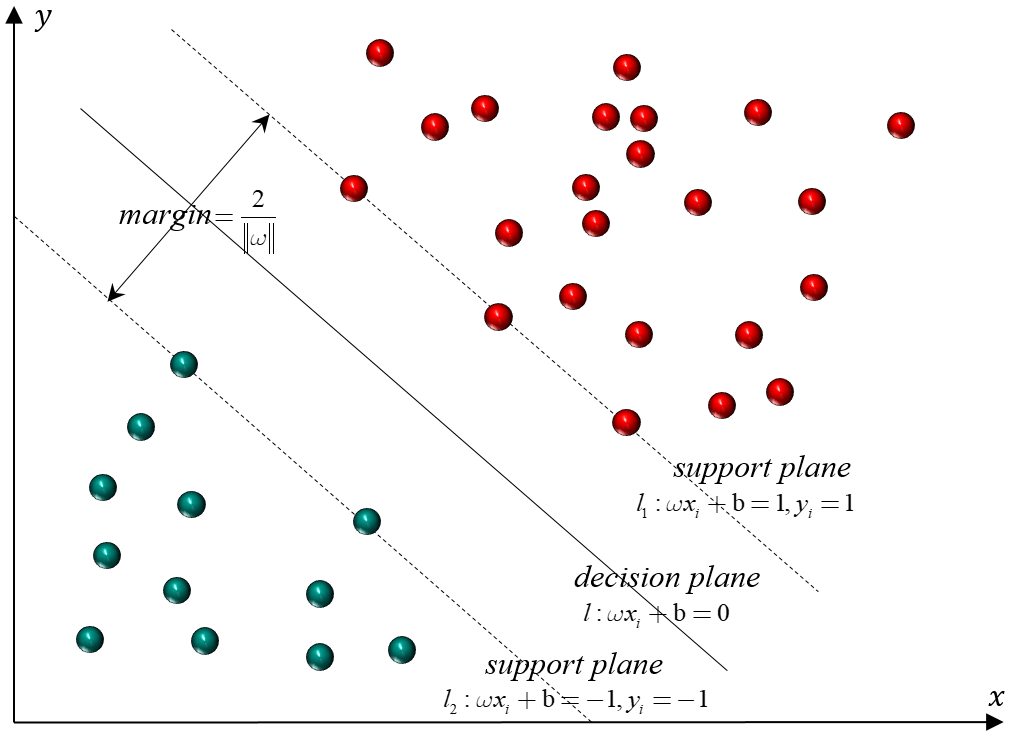}
	\includegraphics[width=1.5in]{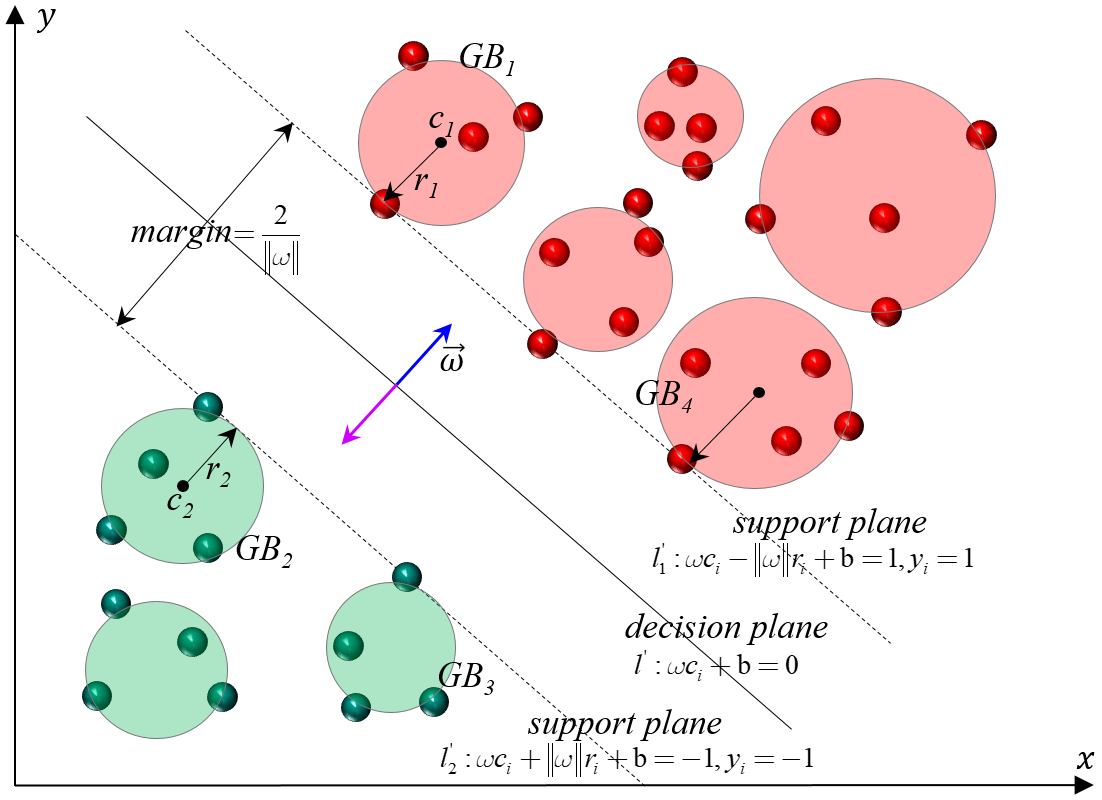}\\
	\centering
	(a)\quad\quad\quad\quad\quad\quad\quad\quad\quad(b)\\
	\centering
	
	\caption{Comparison between traditional SVM and GBSVM~\cite{xia2024gbsvm}. (a) Traditional SVM; (b) Granular-ball SVM.}
	\label{fig:GBSVM}  
\end{figure}

Building on the basic granular-ball support vector machine, Quadir et al. proposed the pinball-loss-based granular-ball twin support vector machine (Pin-GBTSVM)~\cite{GBTSVMTNNLS,quadir2024granular}. This method takes granular balls rather than individual sample points as classifier inputs and enhances noise robustness by introducing the pinball loss. They further extended it to the large-scale setting and developed Pin-LGBTSVM, which avoids matrix inversion to improve computational efficiency and incorporates structural risk minimization through regularization to alleviate overfitting. Experiments on Alzheimer's disease and schizophrenia recognition tasks demonstrated that Pin-LGBTSVM outperforms conventional methods~\cite{quadir2024granular}. Anouck Girard et al. further proposed a multiclass framework, namely the granular-ball $K$-class twin support vector classifier (GB-TWKSVC)~\cite{ganaie2025granular1}. This method improves robustness to noise through granular-ball representation and exploits the nonparallel hyperplane structure of TWSVM to transform multiclass classification into a set of smaller quadratic programming problems, thereby improving computational efficiency. Experimental results show that GB-TWKSVC achieves superior classification accuracy and computational performance on multiple benchmark datasets, indicating its potential in pattern recognition, fault diagnosis, and large-scale data analysis. More recently, Lang et al. proposed the granular-ball fuzzy twin support vector machine~\cite{GBFTSVM2025}, which exhibits strong robustness. In addition, Ganaie et al. introduced universum samples into the TSVM framework together with granular-ball computing and proposed the granular-ball universum twin support vector machine (GBU-TSVM)~\cite{ganaie2025granular}, significantly improving both noise resistance and computational efficiency. Beyond the TSVM family, Sajid et al. developed a granular-ball-based random vector functional link classifier (GB-RVFL)~\cite{sajid2024gb}, which uses granular balls as the inputs to RVFL and yields more robust classification performance.
 
\subsection{Supervised Granular-Ball Neighborhood Relations}

The $k$-nearest neighbor ($k$NN) method enjoys several advantages, including simplicity, online applicability, natural support for multiclass classification, and ease of implementation. However, its performance is highly sensitive to the choice of $k$, and nearest-neighbor search usually requires a large number of point-wise distance computations, resulting in relatively high time complexity. From a multi-granular perspective, a fixed $k$ essentially corresponds to a single-granularity neighborhood relation. To address this limitation, Xia et al.~\cite{xia2022efficient,xia2019granular} introduced granular balls into $k$NN and proposed the granular-ball $k$-nearest neighbor (GB$k$NN) method. Its basic idea is that the label of a query sample is determined by the label of its nearest granular ball, while the label of a granular ball is determined by the majority label of the samples it contains. Compared with traditional $k$NN, GB$k$NN no longer requires explicit optimization of the parameter $k$, but instead performs classification through adaptively generated multi-granular neighborhood relations, thereby improving efficiency, robustness, and neighborhood adaptability.
 
 \subsection{Granular-Ball Rough Sets}

\begin{figure}[htbp!]
 	\centering
 	\setcounter{subfigure}{0}
 	\includegraphics[width=1.55in]{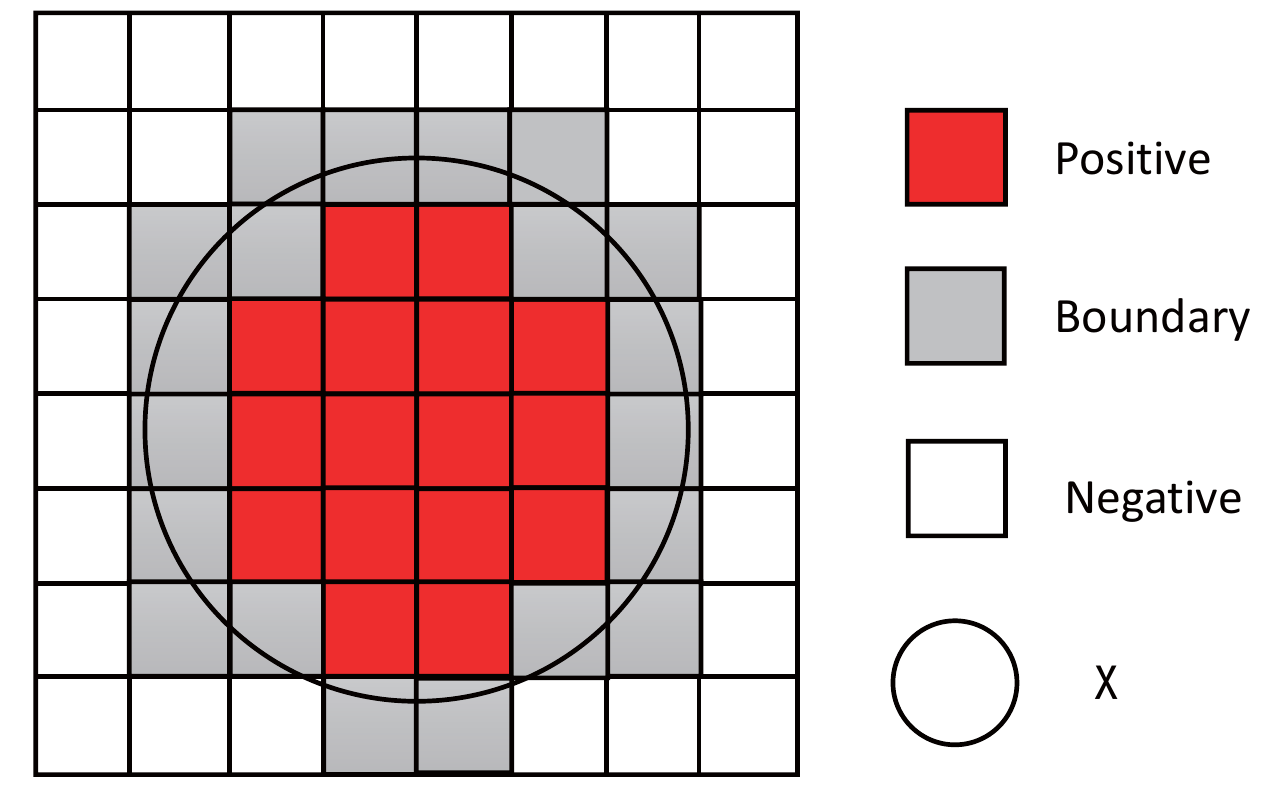}
 	\includegraphics[width=1.55in]{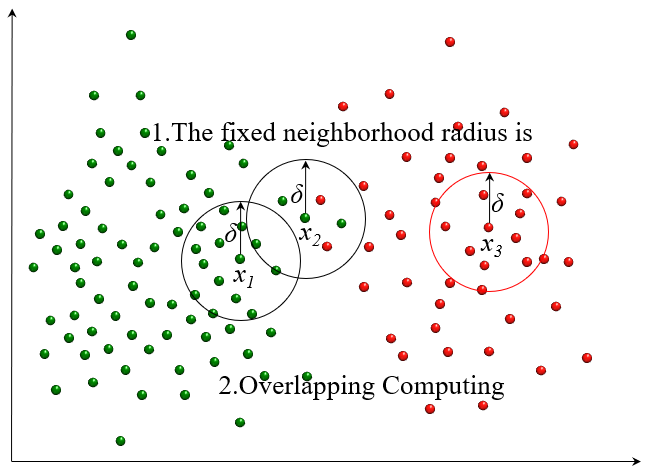}\\
 	\centering
 	(a)\quad\quad\quad\quad\quad\quad\quad\quad\quad(b)\\	
 	\includegraphics[width=1.6in]{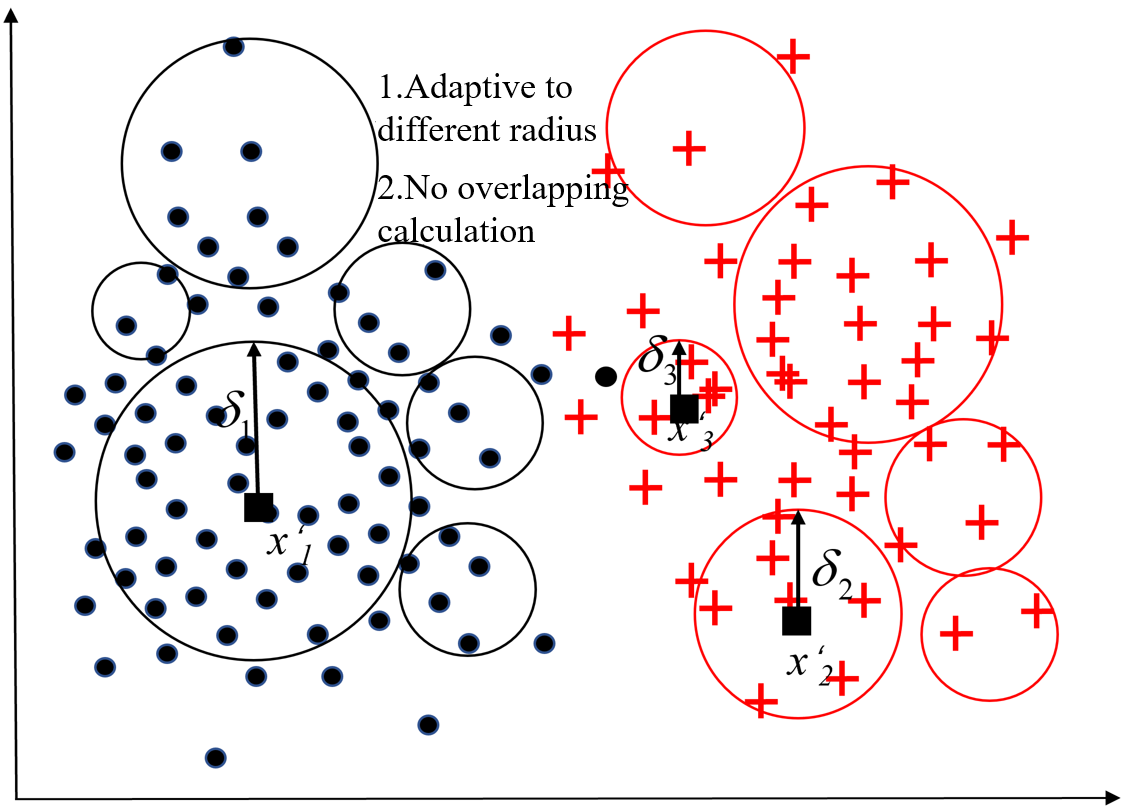}\\
 	\centering
 	(c)\\
 	\caption{The heterogeneous-class propagation phenomenon in traditional neighborhood rough sets~\cite{xia2023gbrs}. (a) Classical rough sets; (b) heterogeneous-class propagation in traditional neighborhood rough sets; (c) granular-ball rough sets eliminate heterogeneous-class propagation.}
 	\label{fig:gbnrs} 
 \end{figure}

 Rough set theory is a fundamental framework in multigranular computing and is widely used in feature selection, classification, and knowledge representation. Classical Pawlak rough sets offer good interpretability through equivalence classes and naturally support multigranular knowledge representation, but they cannot directly handle continuous data. Neighborhood rough sets, in contrast, can process continuous data, yet lose the equivalence-class-based representation of lower and upper approximations~\cite{hu2008neighborhood}. The root cause is that traditional rough-set models still rely on point-wise computation, with neighborhood relations defined at the point level, which has long created a tension between continuous-data processing and knowledge representation. To address this issue, Xia et al.~\cite{xia2023gbrs} identified that existing neighborhood rough sets suffer from the phenomenon of ``heterogeneous transmission,'' as illustrated in Fig.~\ref{fig:gbnrs}(b). As shown in Fig.~\ref{fig:gbnrs}(c), granular-ball computing avoids this problem by removing overlaps between heterogeneous classes and constructing adaptive granular-ball neighborhoods to define indiscernibility relations, equivalence relations, lower and upper approximations, and positive and negative regions. Based on this idea, they established granular-ball rough sets as a unified model of classical rough sets and neighborhood rough sets, thereby improving the capability of rough-set analysis. Early on, Xia et al.~\cite{xia2020gbnrs} proposed granular-ball neighborhood rough sets (GBNRS), which replace the classical positive region in neighborhood rough sets with generated positive regions for attribute reduction. This not only improves neighborhood computation efficiency, but also adaptively determines neighborhood radii consistent with local data distributions. Chen et al.~\cite{chen2025fuzzy} further proposed a variable-precision model integrating granular-ball rough sets, fuzzy rough sets, and neighborhood rough sets, and developed a fuzzy neighborhood granular-ball rough set model. To reduce the high computational cost of GBNRS in high-dimensional spaces, Zhang et al.~\cite{zhang2025attribute} proposed a fast variable granular-ball generation model based on a granular-ball quality index, which employs an adaptive mechanism to optimize radii and efficiently characterize class separability. Zhang et al.~\cite{zhang2023incremental} further introduced an incremental granular-ball neighborhood rough set method. Sun et al.~\cite{SUNGBFNRS} proposed granular-ball fuzzy neighborhood rough sets for high-dimensional multiobjective mayfly-optimization-based feature selection. Ye et al.~\cite{ye2025innovative} developed a new neighborhood-search-based granular-ball generation method that can generate granular balls in multiple granularity spaces; by introducing a generalized multigranular granular-ball rough set model, they reduced the required number of granularity spaces and significantly improved both computational efficiency and feature-selection effectiveness. Zhang et al.~\cite{zhang2025novel} combined granular-ball neighborhood rough set theory with the Wu--Leung model to construct multiscale granular-ball neighborhood decision tables, thereby improving attribute reduction efficiency and classification performance for high-dimensional data. Xia and Lian et al.~\cite{xia2025gbfrs} further incorporated granular-ball computing into fuzzy rough set theory by replacing sample points with granular balls, thus enhancing robustness. In addition, Tan et al.~\cite{tan2025granular} represented groups of similar instances by fuzzy granular balls and introduced a monotonically evolving granular-ball partition mechanism, leading to more stable decision boundaries and more reasonable evaluations of attribute importance.

\subsection{Granular-Ball Concept Cognition}

Concept cognitive learning (CCL) is an important branch of traditional granular computing and is known for its strong knowledge representation ability and interpretability. However, when handling continuous data, it usually relies on fuzzy information granules, and its cognitive operators and adaptive knowledge representation capability remain sensitive to granularity design. To address this issue, Li et al.~\cite{guo2025granular} developed a granular-ball-based concept learning method from the supervised granular-ball representation model. By using ``granular-ball fuzzy concepts'' as the basic units of the concept space, this method jointly encodes sample distributions and label semantics, enabling concept representation in continuous spaces while improving decision efficiency. Furthermore, to overcome the complexity, instability, and limited uncertainty modeling of existing concept cognitive learning methods in complex data scenarios, Guo et al.~\cite{guo2026adaptive} proposed a granular-ball concept cognitive learning framework, which adaptively generates granular balls to capture data at multiple granularities and uses them as the basic information granules for concept cognition. This framework preserves concept interpretability while improving the efficiency, stability, and adaptability of cognitive learning.
 
 \subsection{Granular-Ball Fuzzy Sets}
 
 Fuzzy set theory is a classical framework for uncertainty analysis and decision making, but most existing methods still operate on sample points as the basic computational units, which limits their efficiency and robustness to noise. To address this issue, Xia et al.~\cite{xia2024fuzzy} proposed a granular-ball fuzzy set framework that replaces sample points with granular balls as the basic input units for fuzzy computation. The membership of a granular ball is defined as either the average membership of the samples within the ball or the membership of its center, thereby improving both efficiency and robustness. Based on this idea, incorporating the granular-ball membership $\delta_i$ into granular-ball SVM yields the granular-ball fuzzy SVM (GBFSVM). When the granular-ball radius $r_i=0$, GBFSVM reduces to the traditional fuzzy SVM; in the general case, it combines the advantages of fuzzy weighting and granular-ball representation. In addition, the granular-ball paradigm has been further extended to related studies such as fuzzy feature selection and fuzzy spatial reasoning. Qian et al.~\cite{QIAN2025122047} proposed a granular-ball-based method for fuzzy relevance and redundancy analysis, which uses granular-ball sets to characterize the supervised information of candidate and non-candidate labels, and combines granular-ball consistency with fuzzy entropy to construct a new fuzzy mutual information measure for feature ranking under a relevance--redundancy trade-off. Kwon et al.~\cite{kwon2025fgb} proposed the FGB-OPRAm framework, which integrates fuzzy granular balls with OPRAm-based spatial reasoning to provide a more flexible and reliable solution for spatial queries in uncertain environments.

\subsection{Granular-Ball Continual Open Learning}

In continual and open environments, feature selection and class recognition must simultaneously address challenges such as knowledge transfer, unknown-class detection, and dynamic updating. To tackle this problem, Yang et al.~\cite{cao2024open} proposed the granular-ball-based continual open feature selection framework, which uses multi-granular granular-ball representations to construct and update a knowledge base, thereby supporting unknown-class identification, old-knowledge transfer, and feature-subset enhancement during open learning. Furthermore, Li et al.~\cite{li2025multi} proposed a multigranular open intent classification method, which distinguishes known and unknown classes more effectively through adaptive granular-ball clustering and multi-granular decision-boundary modeling. These studies show that granular-ball computing can enhance adaptability and generalization in open environments through multi-granular knowledge representation, dynamic knowledge updating, and adaptive boundary modeling.
 
 \subsection{Other Granular-Ball Classification Learning Methods}
 For partial-label learning (PLL), which is a challenging weakly supervised multiclass problem often accompanied by class imbalance, existing methods typically rely only on inter-class pseudo-label information and overlook the combined effect of intra-class imbalance and inter-class features. To address this limitation, Cheung et al.~\cite{GBRIP2025} proposed the granular-ball representation for imbalanced partial-label learning (GBRIP) together with a corresponding learning framework. This method constructs the feature space using coarse-grained granular-ball representations and introduces a multi-center loss to better capture intra-class distributions, thereby reducing the influence of confusing features in both label disambiguation and imbalanced-distribution modeling. Extensive experiments on standard benchmarks show that GBRIP outperforms existing advanced methods and provides a robust solution to imbalanced PLL. To reduce secondary errors in partial-label multiclass learning, Sun et al.~\cite{SUNFGBCPMFS} proposed a fuzzy mutual-information-based partial-label feature selection method built upon fuzzy granular-ball clustering. For partial multi-label feature selection (PMLFS), where candidate label sets often contain false-positive labels and existing methods rely excessively on topological information and linear regression models while failing to characterize local structures, Qian et al.~\cite{xu2025granular} proposed a two-stage PMLFS method based on granular-ball computing. In the first stage, granular-ball computing is used to model sample distributions under different labels and perform label disambiguation; in the second stage, fuzzy decision neighborhood rough sets are employed to exploit local sample structures, maximize the consistency of relevant labels, and suppress the uncertainty introduced by irrelevant ones, thereby significantly improving both effectiveness and robustness. Granular-ball computing has also been applied to sampling and robust learning in large-scale classification. To handle nonlinear classification with low time complexity, Xia et al.~\cite{xia2021granular} used granular balls to fit classification data and selected several points closest to the coordinate-axis directions on the ball boundary as representative samples. Xie et al.~\cite{GBS2025} further addressed the boundary-overlap problem in granular-ball sampling, thereby improving learning accuracy. They also proposed a granular-ball AdaBoost method by combining granular-ball-based data compression with the AdaBoost classifier, which improves both robustness and efficiency over traditional AdaBoost~\cite{GAdaBoost2025}. In noisy-label scenarios such as industrial fault diagnosis, Deng et al.~\cite{MgCNL2025} proposed a robust learning method based on granular-ball confidence evaluation. By combining contrastive learning with multi-granular granular-ball modeling, the method estimates sample confidence, selects high-confidence samples for training, and adaptively adjusts granular-ball granularity, thereby alleviating the effect of noisy labels.

\section{Granular-Ball Unsupervised Learning}

Unsupervised learning aims to uncover the intrinsic structure of data in the absence of label information, among which clustering is one of the most representative tasks. Existing clustering methods generally rely on point-wise computation or single-granularity relations and therefore often suffer from limitations in efficiency, robustness, and generalization. To address these issues, granular-ball clustering adaptively and efficiently generates granular-ball representations, replacing the point-wise inputs used in traditional clustering, and has led to a series of clustering models built upon such representations.

\subsection{Unsupervised Granular-Ball Representation Model}

In unsupervised learning, granular balls are still typically represented as standard spheres, i.e., $GB_i=(X_i,c_i,r_i)$. By jointly optimizing the centers $c_i$, radii $r_i$, and the number $m$ of granular balls, the following unsupervised granular-ball representation model can be formulated:
\begin{equation}\label{equ:GBC_clustering}
	\begin{aligned}
		& \min_{c_i,r_i,m} \ \mathcal{J} (-Cov(GB),-Comp(GB)) + m \\
		& \text{s.t.} \quad quality(GB_i) \geq  
		\frac{|GB_{i1}|}{|GB_i|} \cdot quality(GB_{i1})+ \\
		&\frac{|GB_{i2}|}{|GB_i|} \cdot quality(GB_{i2})
		+
		\cdots
		+
		\frac{|GB_{ik}|}{|GB_i|} \cdot quality (GB_{ik}),\\
		& quality(GB_i)= \frac{1}{|GB_i|}  \sum_{ x \in GB_i}  \| x- c_i\|, \ i = 1,2,\dots,m,
	\end{aligned}
\end{equation}
where the coverage term $Cov(GB)$ reflects how completely the granular-ball set represents the original data, while the compactness term $Comp(GB)$ characterizes the concentration of samples within each granular ball. A more compact distribution is more favorable for representing intra-cluster structures and inter-cluster differences~\cite{GBCTPrincipleJG2025}. In unsupervised settings, granular-ball quality is usually measured by the average distance from the samples within the ball to its center~\cite{GBCT2024}; the smaller this value is, the more compact and higher-quality the granular ball becomes. In practice, unsupervised granular-ball generation typically adopts full coverage and enforces compactness through radius control and adaptive splitting. Its central goal is therefore to achieve an effective representation of the data distribution with as few granular balls as possible under the requirement of quality improvement. Correspondingly, whether a granular ball should be further split is usually determined by whether the weighted sum of the qualities of the resulting child granular balls is improved, thus balancing representation accuracy and computational efficiency.

\subsection{Generation and Optimization of Unsupervised Granular-Ball Representation Models}

\subsubsection{Basic Model of Granular-Ball Generation}

\begin{figure}[htbp!]
	\centering
	\includegraphics[width = 1\linewidth]{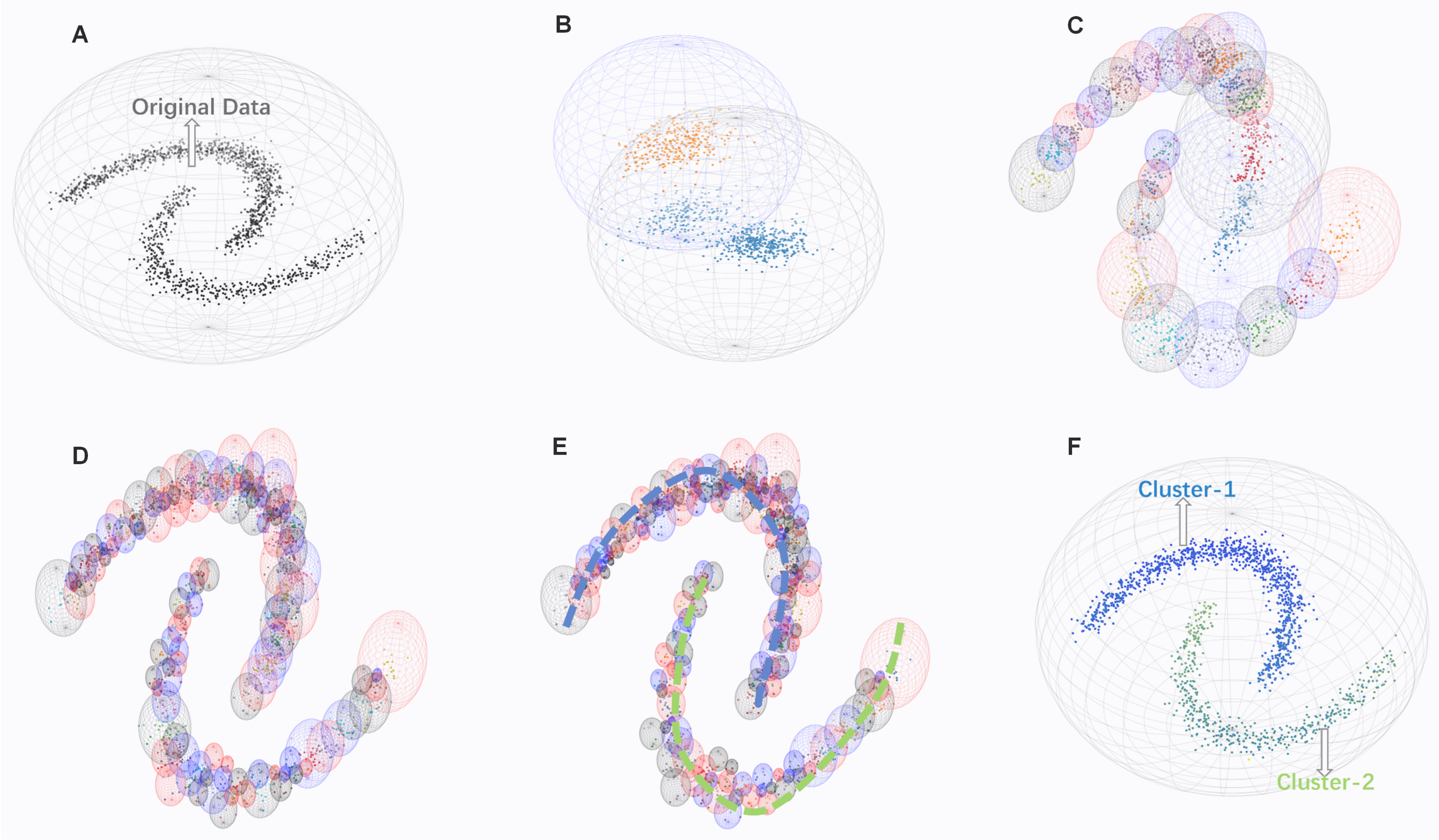}
	\caption{Schematic diagram of the basic granular ball clustering algorithm~\cite{GBMGNR2024}.}
	\label{fig:renzhi1}
\end{figure}

The generation and optimization of unsupervised granular-ball representations are broadly similar to those in supervised settings, with the main differences lying in the quality measure and the splitting criterion. Existing unsupervised granular-ball generation methods typically adopt a binary-splitting strategy, in which each granular ball is divided into two child balls, and two relatively distant samples are selected as the initial centers to obtain a better split. Granular-ball quality is usually measured by the average distance from the samples within the ball to its center, where a smaller value indicates a more compact granular ball; splitting continues if the weighted quality of the child balls is better than that of the parent ball, and otherwise stops~\cite{GBCT2024}. Fig.~\ref{fig:renzhi1} illustrates the basic granular-ball generation algorithm. Starting from a single large granular ball covering the dataset, the method progressively evolves from coarse to fine through binary splitting, and determines whether to continue splitting according to the quality of the parent and child balls. Under the full-coverage constraint, this approach improves compactness while maintaining a relatively small number of granular balls, thereby achieving a multi-granular representation. As a result, more and smaller granular balls are generated in complex regions, whereas fewer and larger ones are produced in simple regions; noisy samples tend to form very small or isolated granular balls, naturally capturing anomalies and boundary points. Since it does not require computing all pairwise distances, its overall complexity is approximately $O(nm)$, making it more efficient, robust, and adaptive than traditional clustering methods.

\subsubsection{Granular-Ball Generation From the Perspective of Justifiable Granularity}

To address the over-reliance of traditional granular-ball generation methods on a single quality measure and thresholding strategy, Jia et al.~\cite{GBCTPrincipleJG2025} proposed GB-POJG, a granular-ball generation method grounded in the principle of justifiable granularity. The method characterizes granular-ball quality in a unified manner from two competing perspectives, namely coverage and specificity, and seeks to maximize the overall quality of the granular-ball set. By combining a granular-ball binary tree with tree-based dynamic programming, it automatically determines whether a split should be retained or pruned, thereby producing a granular-ball set that better matches the underlying data distribution. In addition, it introduces abnormal granular-ball detection and progressive refinement to correct unreasonable structures. GB-POJG achieves a unified optimization of coverage, compactness, and overall quality, and thus provides a more stable and principled solution for granular-ball construction in unsupervised clustering.

\subsection{Clustering Models Based on Granular-Ball Representations}

The basic granular-ball clustering method can serve not only as an independent clustering algorithm, but also as an efficient and adaptive multi-granular representation mechanism for improving existing clustering models. Its key advantage lies in replacing point-wise inputs with granular balls, thereby alleviating the issues of fine-grained computation, parameter sensitivity, and limited robustness in traditional clustering. Based on this idea, a variety of granular-ball clustering models have been developed under different clustering paradigms~\cite{GBCT2024}.

To address the $O(n^2)$ time complexity of traditional density peaks clustering (DPC), which requires computing a point-wise pairwise distance matrix, Cheng et al.~\cite{10215064} proposed the granular-ball density peaks method GB-DP. This method first replaces sample points with granular balls, then defines density and relative distance at the granular-ball level, and constructs a granular-ball decision graph for cluster-center selection and label propagation, thereby significantly reducing the time and space costs in large-scale settings. Furthermore, to overcome the topology distortion often introduced by existing acceleration strategies for density-based clustering, Chen et al.~\cite{chen2025gbsk} proposed the granular-ball skeleton clustering (GBSK) method. Built upon modal clustering and kernel density estimation, GBSK approximates local density order through granular-ball density, and progressively reconstructs the geometric density skeleton of the data via repeated sampling, representative granular-ball identification, and key granular-ball refinement, thereby achieving near-linear or even linear complexity while preserving topological fidelity.

To address the large point-wise similarity matrix and noise sensitivity in spectral clustering, Xie et al.~\cite{xie2023efficient} introduced granular-ball computing into spectral clustering. By replacing sample points with granular balls when constructing the similarity matrix, the method significantly reduces the size of the adjacency matrix and improves both the rationality and robustness of similarity construction by defining granular-ball similarity based on the relations among centers and radii. Cheng et al.~\cite{GBCPLT2025} further proposed a granular-ball pseudo-label spectral clustering method, which first generates pseudo-labels by weighted $k$-means and refines the representation using granular balls, then uses granular-ball centers as high-quality anchors to construct a sample--anchor similarity matrix, and finally performs fast eigendecomposition via transfer cut, thereby improving both efficiency and accuracy in high-dimensional spectral clustering.

To address the limited efficiency and noise sensitivity of traditional multiple-kernel $k$-means, which relies on point-to-point optimization under complex distributions, Xia et al.~\cite{GBMKeans2025} proposed the granular-ball kernel $k$-means clustering method. This method replaces sample points with granular balls in multiple kernel spaces, defines granular-ball kernels, and constructs a granular-ball-based multiple-kernel optimization model. As a result, the modeling granularity of multiple-kernel clustering is elevated from the point level to the granular-ball level, which enhances the characterization of meso-scale structures and nonlinear relations while alleviating the computational burden and noise interference caused by kernel interactions.

For graph neural network clustering on high-dimensional non-graph data, where irrelevant features are highly disruptive and graph construction often depends on manually tuned parameters, Xie et al.~\cite{xie2025tpami} proposed the adaptive weighted granular-ball graph autoencoder clustering method. This method first weakens the effect of irrelevant features through feature weighting and generates a weighted granular-ball set in a bottom-up manner; it then constructs a hierarchical graph structure by combining local connections within granular balls and global connections between granular balls. A dynamic stepwise update mechanism is further introduced so that the adjacency range can be adaptively adjusted during training, thereby unifying feature selection, granular-ball representation, graph-structure learning, and graph neural network clustering within a single framework.

\subsection{Other Granular-Ball Clustering Models}

Beyond the representative methods discussed above, granular-ball ideas have also been widely used to improve a variety of clustering models. In minimum spanning tree (MST) clustering, Xie et al.~\cite{xie2023gbmst} replaced sample points with granular balls to construct the MST, thereby reducing the scale of node and edge-weight computations and improving both efficiency and robustness over traditional point-wise MST clustering. In stream clustering, Wang et al.~\cite{GBFuzzyStream2024} proposed a granular-ball streaming clustering method, which uses granular balls as the unified representation units in both online and offline stages and incorporates a decay mechanism to respond efficiently and adaptively to data drift. For clustering based on neighborhood relations, Xie et al.~\cite{GBMGNR2024} proposed a multigranular neighborhood relation based on granular balls, and applied it to both $k$-nearest neighbor classification and DBSCAN clustering, achieving improved learning accuracy. Granular-ball computing has also been integrated into density-based clustering. NaGB-DBSCAN combines natural neighborhoods with granular-ball representations, reducing computational cost while lowering the time complexity to $O(N\log N)$~\cite{luo2025nagb}. $GBK$-DPC further improves boundary resolution and fault tolerance through adaptive granular-ball optimization and dual-factor density evaluation~\cite{ni2025gbk}. In multi-view clustering, Su et al.~\cite{MGBCC2025} constructed coarse-grained granular balls and exploited both intra-view and inter-view relations to perform multigranular contrastive learning, thereby better preserving local topological structures. Liu et al.~\cite{VAF2026} modeled both shared and complementary representations by granular balls and further enhanced the discriminability and robustness of multi-view representations through an attention-based fusion mechanism. In addition, in weighted clustering and fuzzy $c$-means clustering, granular-ball methods can improve adaptability to high-dimensional data and complex distributions through local feature weighting and granular-ball-based membership iteration~\cite{xie2023research,WGBC2024}. Granular balls have also been used to accelerate exact $k$-means by serving as cluster-level compressed representations, which enable the discovery of coarse-grained inter-cluster neighborhood relations and reduce distance computations from sample points to non-neighboring clusters, thus significantly improving the efficiency of exact $k$-means, especially when $k$ is large~\cite{xia2020ball}.

\subsection{Outlier Detection via Granular-Ball Computing}

In recent years, granular-ball computing has also been applied to unsupervised tasks such as outlier detection, where robustness and efficiency are of primary concern. To address the coexistence of local and collective outliers, for which a single-scale detection strategy is often insufficient, Gao et al.~\cite{GBODWiltold2025} proposed the multiscale granular-ball outlier detection method (MGBOD). By jointly modeling fine- and coarse-grained views and combining relative fuzzy granularity density, multiscale fusion, and three-way decision, MGBOD provides a unified framework for outlier scoring and decision making. Beyond multiscale methods, granular-ball representations have also been used to improve mean-shift-based and local-density-based outlier detection. Cheng et al.~\cite{cheng2025gbmod} performed mean-shift updates using granular-ball centers as anchors, thereby elevating point-wise nearest-neighbor search to structure-level search. Su et al.~\cite{GBDO2025} constructed local granular-ball densities based on fuzzy similarity relations among granular balls, in order to estimate relative sparsity and derive outlier factors. These methods demonstrate the ability of granular-ball computing to balance compressed representation with the preservation of local topology and density variation, thereby jointly improving the efficiency, robustness, and interpretability of outlier detection.

\section{Approximate Granular-Ball Computing Models}

The standard spherical granular ball serves as the basic representation form in granular-ball computing because of its simplicity and scalability in high-dimensional spaces. However, in low-dimensional settings and task-specific scenarios, a single spherical structure may be insufficient to capture complex local geometric distributions. To address this limitation, approximate granular-ball representations such as granular rectangles, square granular balls, and granular ellipsoids have been introduced to improve the modeling of regular structures, directional distributions, and non-uniform regions.

\subsection{Rectangular Approximate Granular-Ball Representation for Images}

In the visual domain, since pixels are naturally organized in rectangular spatial layouts and neighborhood-based similarity is usually not strictly Euclidean, it is more appropriate to adopt a non-standard granular-ball representation based on rectangular approximation~\cite{xia2025graphRespresation}, namely the granular rectangle. In this case, $\vec{\theta_i}$ consists of the rectangle height $h_i$ and width $w_i$, and the representation is written as $GB_i=(X_i,c_i,h_i,w_i)$. Although it differs from the standard spherical form, the representation of granular rectangles, the coarse-grained-first optimization process, and the subsequent computational framework all strictly follow the principles of granular-ball computing. Therefore, in most cases, ``granular ball'' and ``granular rectangle'' need not be distinguished rigidly.

Since coverage is fixed to full coverage in image granular-ball representation, i.e., $Cov(GB)=1$, and compactness can be satisfied through rectangular coverage, the model mainly optimizes the centers, sizes, and number of granular rectangles. Based on the general granular-ball model (\ref{equ:GBC_1}), the following image granular-ball representation model can be established:
\begin{equation}\label{equ:GB_image}
	\begin{aligned}
		&\min_{c_i,h_i,w_i,m} \ m \\
		& s.t. \quad quality(GB_i) \geq \phi(X), \ i=1,2,\ldots,m, \\
		& quality(GB_i)=\frac{1}{|GB_i|}\sum_{x\in GB_i}\mathbb{I}\big(|f(x)-f(c_i)|<thr\big).
	\end{aligned}
\end{equation}
Here, $|GB_i|$ denotes the number of pixels in region $GB_i$, $f(x)$ denotes the gray value, $thr$ is the threshold for identifying \emph{normal} pixels, and $\mathbb{I}(\cdot)$ is the indicator function. $\phi(X)$ can be specified as a constant threshold or determined adaptively according to the overall noise level or gradient statistics of the image. This model improves the internal representation quality of granular balls through compactness constraints, while achieving a coarser global representation by minimizing the number of granular rectangles.

\subsection{Optimization Algorithm for Image Granular-Ball Representation}

\begin{figure}[!ht]
\centering
\includegraphics[width = 1\linewidth]{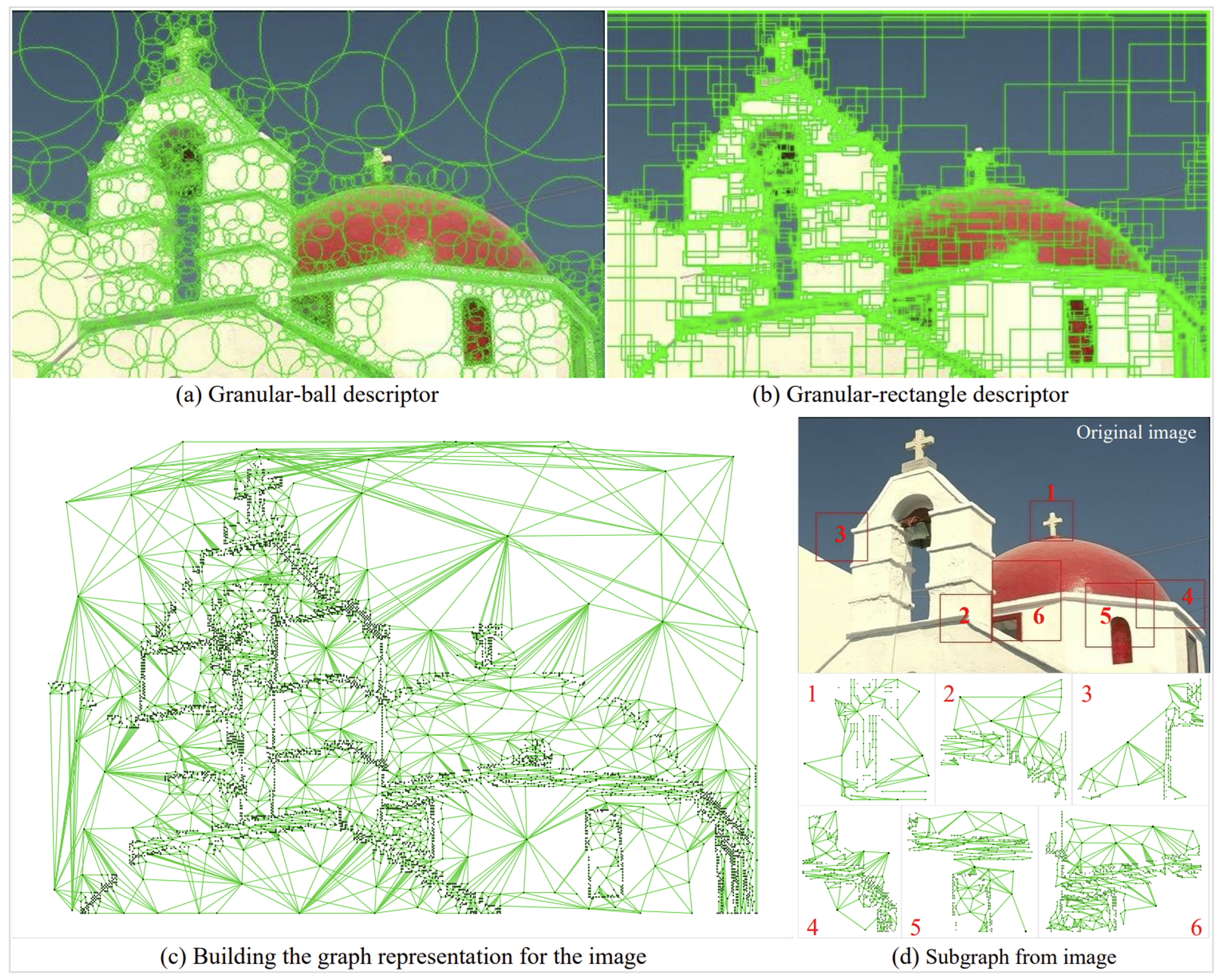}	
\caption{Granular-ball image representation diagram~\cite{xia2025graphRespresation}.}
\label{fig:GBIR}
\end{figure}

The generation and optimization of image granular-ball representations also follow the principles of global-first and coarse-grained priority. Since an image matrix can be viewed as a graph structure with naturally defined pixel adjacency, a sequential generation strategy is adopted rather than a splitting-based one. Specifically, the image is first smoothed with a Gaussian filter and its gradient is computed. A position with the minimum gradient in the uncovered region is then selected as the center of a granular rectangle, which is subsequently expanded along the horizontal and vertical directions until the region purity satisfies the threshold or the minimum size is reached. 
Smooth regions thus form large rectangles, whereas boundaries and complex textures are represented by smaller ones. 
Furthermore, each granular rectangle can be regarded as a node in a graph, with edges established between spatially adjacent or overlapping regions, and node attributes described by low-dimensional statistics such as position, size, color, and texture. This yields a multi-granular graph representation that provides a basis for subsequent graph learning. As shown in Fig.~\ref{fig:GBIR}, this sparse multi-granular graph preserves salient structures and boundaries while reducing representation complexity~\cite{xia2025graphRespresation}.


\subsection{Computational Models Based on Image Granular-Ball Graph Representations}

Although convolutional neural networks (CNNs) exhibit a certain degree of hierarchical semantics through their multilayer architectures, their multi-granularity remains incomplete. On the one hand, convolution kernels are usually of fixed size and therefore cannot adaptively reflect scale variations across different image regions; on the other hand, CNNs still take the finest-grained pixels as input, which naturally limits their robustness and interpretability. To address this issue, the image granular-ball representation method~\cite{xia2025graphRespresation} adaptively extracts multi-granular structural blocks and their relations from an image, and further constructs a granular-ball graph representation. Combined with graph neural models such as GAT, this representation enables visual learning through node feature aggregation, global pooling, and classification, thereby directly incorporating the multi-granular structural information of images into the computation process and improving the efficiency, robustness, and interpretability of visual models.

In addition, Liu et al.~\cite{GBFaexp2025} proposed the CS-GBSBF framework for facial expression recognition, whose core idea is to explicitly construct a multi-granular spatial graph at the raw-image level using granular balls and to use it to guide visual feature fusion. The method first adaptively partitions a facial image into a set of rectangular granular-ball regions and constructs the corresponding graph structure. It then extracts image features and graph-structure features through a visual branch and a spatial branch, respectively, and aligns them via an attention mechanism. By further combining component-level representations with a bootstrap alignment loss, CS-GBSBF achieves stable fusion of spatial and visual features, thereby improving facial expression recognition performance.

\subsection{Parallelizable Square Granular-Ball Computing Models}

From the perspective of superpixel research, existing methods can be broadly divided into traditional non-end-to-end approaches and trainable end-to-end approaches, while the latter usually require explicit modeling of pixel--superpixel assignment relations and additional constraints such as compactness and connectivity, leading to high implementation complexity and memory overhead. In contrast, square granular-ball image representation dynamically adjusts block sizes on a regular grid through purity-driven selection and refinement, allowing homogeneous regions to remain large while progressively refining complex regions, thereby providing stable, controllable, and efficient regional units~\cite{xia2026squaresuperpixelgenerationrepresentation,zhao2025square}. Since these blocks are naturally aligned with the grid, they can be directly tensorized and support token pruning, which reduces the computational cost of attention. Compared with methods based on irregular superpixels, square granular-ball blocks maintain regular alignment and a fixed budget while adaptively allocating granularity, making them more suitable for graph construction, message passing, and attention aggregation. As such, they provide a parallel-friendly multi-granular representation foundation for Transformers, multimodal alignment, and graph-based visual learning.

 \subsection{Computational Models Based on Granular Ellipsoids}
 
 The standard spherical granular ball is the most basic and widely used representation in granular-ball computing, owing to its simple, unified, and symmetric mathematical form as well as its good scalability in high-dimensional spaces. However, when data in low-dimensional spaces exhibit strong directionality, scale inconsistency, or anisotropic distributions, a single spherical structure is often insufficient to capture local geometric characteristics. To address this limitation, researchers have further extended granular-ball representation to granular ellipsoids. Granular ellipsoids follow the same basic principle of replacing point-wise inputs with meso-level structural units, while generalizing the geometry from spheres to ellipsoids. Compared with standard granular balls, granular ellipsoids describe local regions not only by their centers but also by additional parameters such as axis lengths, orientations, or covariance, thereby providing a finer characterization of nonspherical local structures and feature correlations. It should be noted that granular ellipsoid representation is mainly suitable for low-dimensional scenarios, since in high-dimensional spaces the parameterization and estimation of ellipsoids become substantially more complex, weakening their modeling and computational advantages.
 
 In supervised learning, Sun et al.~\cite{sun2025gec} replaced traditional point-wise samples with granular ellipsoids and modeled local samples by minimum-volume enclosing ellipsoids. By jointly capturing location, scale variation, and directional information, their method not only inherits the efficiency and robustness brought by replacing point-wise samples with meso-level granular units, but also shows favorable classification efficiency and discriminative ability when handling irregular class distributions and complex local structures. In unsupervised learning, Guo et al.~\cite{guo2026granular} used ellipsoidal local sample regions in density peaks clustering. Because ellipsoidal granular structures are more suitable for characterizing directionality, nonuniformity, and local anisotropy in complex data, they can better adapt to complex data distributions while preserving the computational advantages of granular structures, thereby improving clustering efficiency and robustness. Liu et al.~\cite{liu2026fast} further proposed a fast granular-ellipsoid density peaks clustering algorithm for large-scale data, which defines density and relative distance at the granular-ellipsoid level and achieves efficient clustering of nonspherical clusters in large-scale settings.
 
 Kerbl et al.~\cite{kerbl20233d} proposed 3D Gaussian Splatting (3DGS) in 2023, which represents continuous 3D scenes as a set of optimizable 3D Gaussian primitives, i.e., granular ellipsoids. This method reflects three core ideas of granular-ball computing. First, it adopts coarse initialization followed by adaptive refinement, where oversized Gaussians are progressively split into smaller ones, mirroring the coarse-to-fine splitting principle of granular-ball generation~\cite{xia2019granular}. Second, splitting is determined by scale and gradient thresholds, which is analogous to the quality-driven refinement criterion in granular-ball computing~\cite{xia2025graphRespresation}. Third, by controlling Gaussian overlap and pruning redundant structures, it achieves an adaptive sparse-to-dense spatial arrangement, which is close in spirit to the structural control and redundancy pruning strategies in granular-ball representations~\cite{GBCTPrincipleJG2025,xia2025graphRespresation}.

 \section{Granular-Ball Deep Learning Based on Latent-Space Granulation}
 
 In deep learning, granular balls are no longer restricted to explicit construction in the original data space, but are instead adaptively generated in the latent space of neural networks. Granular balls formed in the latent space provide a relatively stable intermediate representation for training data, reducing sample-level variation and mitigating task-level prediction error. When jointly optimized with feature learning, they provide end-to-end models with more stable and robust multi-granular structural representations. Unlike approaches that operate directly in the original data space, existing granular-ball deep learning methods typically generate a set of granular balls $GB=\{GB_i\}_{i=1}^m$ in the latent space $\varphi(X)$, and use them as the basic input units for subsequent learning modules, thereby alleviating sample isolation and distribution imbalance in high-dimensional representation spaces. In general, a granular ball can be written as $
 GB_i=\bigl(\varphi(X_i),c_i,\vec{\theta_i}\bigr)$, where $\vec{\theta_i}$ includes attributes such as radius, scale, or shape. In deep learning tasks, samples (or tokens) usually need to be modeled under full coverage, and thus the coverage term satisfies $Cov(GB)=1$. Based on the general granular-ball model (\ref{equ:GBC}), the deep representation model based on latent-space granulation can be formulated as
 \begin{equation}\label{equ:GBC_deep}
 	\begin{aligned}
 		&\quad \quad \quad \quad f( \varphi(X) ,\vec{\alpha}) \longrightarrow  g(GB,\vec{\beta}) \\
 		&\min_{c_i, \vec{\theta_i},\vec{\beta},m} \ -Comp(GB) + m \\
 		& s.t. \quad quality(GB_i) \geq \phi(\varphi(X)), \ i = 1,2,\ldots,m.
 	\end{aligned}
 \end{equation}
 
 This model improves the internal quality of granular-ball representations through a compactness constraint, while encouraging a coarser global representation by minimizing the number of granular balls. The definition of granular-ball quality $quality(GB_i)$ may vary across tasks. For example, in image classification and natural language processing, it is often characterized by the proportion of dominant labels within a granular ball; in reinforcement learning, it can be defined in terms of the relation between granular-ball capacity and the minimum number of samples; and in time-series analysis, it may be measured by the ratio between the number of samples and the total radius. Accordingly, the splitting constraint function $\phi(\varphi(X))$ can either be specified as a constant threshold or defined adaptively according to the quality change before and after splitting. As the network parameters $\vec{\alpha}$ and task parameters $\vec{\beta}$ are updated during training, the hidden representations $\varphi(X)$ continue to evolve, and the centers, radii, and other attributes of granular balls are adjusted accordingly. Therefore, this framework essentially corresponds to a joint optimization process of \emph{representation learning--granular-ball generation--task learning}, and provides a unified modeling form for building stable and robust deep models in multi-task scenarios.

 \subsection{Image Classification}
 To address the issue of label noise in image classification, Dai et al. proposed a granular-ball computing module (GBC) that can be embedded into CNNs~\cite{dai2024granular}. The method adaptively partitions the hidden representations of a mini-batch in feature space into granular balls, uses the granular-ball centers as new structured representations, and takes the majority label within each granular ball as the supervisory signal, thereby elevating traditional point-wise supervision to multi-granular supervision based on local consensus. Furthermore, by introducing a gradient backpropagation mechanism matched to the granular-ball structure and granular-ball-level experience replay, the method improves the robustness, stability, and generalization of image classification under label noise.

 \subsection{Granular-Ball Reinforcement Learning}
 
 A policy learned in a simple environment can often be transferred to more complex environments that share the same task logic through limited feature-alignment training, a process referred to as cognitive generalization or systematic generalization. To address the insufficient generalization ability of deep reinforcement learning in complex environments, Liu et al. proposed a general granular-ball reinforcement learning framework (GBRL) to improve the cognitive generalization capability of standard DRL methods~\cite{liu2024unlock}. The framework first constructs a cognitive latent space in a simple environment and then partitions it, so that samples with similar environmental effects are represented by granular balls in the same subregion. During fine-tuning in a complex environment, the policy aligns new samples in the cognitive latent space with samples collected from the corresponding subregions in the simple environment, and approximates their rewards and $Q$-values for policy updating. Under the same task logic, GBRL enables effective policy generalization across a variety of challenging environments.
 
 \subsection{Granular-Ball Natural Language Processing}
 
 The advantages of granular-ball computing in natural language processing are mainly reflected in robust learning, particularly for label-noise handling and adversarial defense. To address the inevitable labeling errors in large-scale text annotation, Wang et al.~\cite{GBRAIN2024} proposed the granular-ball-based training framework GBRAIN. By integrating dynamic granular-ball clustering with neural network training, GBRAIN adaptively groups semantically similar samples into the same granular ball and uses the granular-ball center vector and granular-ball label to form coarse-grained representations of the samples within the ball. In this way, the influence of noisy labels is alleviated through majority-correct labeling, while a specially designed backpropagation mechanism is used to continuously optimize the coarse-grained embedding representations. 
 
 To defend against text adversarial attacks based on synonym substitution, Wang et al. further proposed the granular-ball sample enhancement defense framework GBSEF~\cite{GBTAD2024}. This method adaptively granularizes synonym sets through supervised granular-ball representations, replaces original word representations with granular-ball center vectors, and constructs robust text samples that preserve both syntactic and semantic information. It further combines a random substitution mechanism to generate more effective augmented samples for training. Experimental results show that such methods exhibit strong robustness and transferability across multiple neural architectures and attack settings, suggesting that granular-ball computing can provide effective support for robust learning in natural language processing through coarse-grained semantic aggregation and local consistency modeling.
 \subsection{Granular-Ball Time-Series Analysis}
 
 With the development of the Internet of Things and industrial monitoring, time-series anomaly detection faces challenges such as strong continuity, dynamic evolution, multi-density distributions, and noise interference. Traditional methods based on neighborhood relations, clustering, one-class classification, or prototype memory typically rely on fixed structures to describe normal patterns, and are therefore inadequate for adaptively characterizing the flexible and continuous topology of time-series data in feature space. To address this issue, Shen et al.~\cite{Shen2026} proposed a granular-ball one-class network, which introduces a data-adaptive representation called granular-ball vector data description. It partitions the latent space into compact and high-density regions represented by granular balls. These granular balls are generated through a density-guided hierarchical partitioning process and further refined by removing noisy structures, thereby achieving both robustness and efficiency in time-series anomaly detection.

\subsection{Granular-Ball Object Detection}

With the widespread success of deep learning in image classification and object detection, improving model generalization to unseen target domains has become an important research problem. Existing domain adaptation methods usually rely on domain-adversarial learning for feature alignment, but often overlook the influence of latent noncausal factors such as background, illumination, and color, thereby limiting cross-domain transferability. To address this issue, Zhang et al. proposed GB-DAL, a granular-ball-based domain generalization method for object detection~\cite{zhang2026implicit}. The method introduces a granular-ball partition module together with simulated noncausal-factor data augmentation. It first partitions source-domain samples into multiple granular balls through prototype-driven granular-ball splitting and uses granular-ball center vectors for feature representation. It then combines adversarial training with multi-source joint learning to suppress noncausal disturbances and improve cross-domain feature alignment, thereby significantly enhancing the stability and generalization of object detection models in complex environments and unseen target domains.

\subsection{Granular-Ball Domain Generalization for Crowd Counting}

In crowd counting, variations in camera viewpoint, scene layout, object density, and background texture often make models trained on a single source domain difficult to generalize to unseen scenes. To address this issue, Chen et al. proposed a granular-ball-guided stable latent-domain discovery method~\cite{chen2026granular}. The method first constructs granular-ball representations in the latent space and then uses granular-ball centers as more stable meso-level structural units for higher-level clustering. In this way, it reduces the sensitivity of pseudo-domain partitioning to outliers, noise perturbations, and feature fluctuations, thereby improving cross-domain robustness and stability in single-source domain generalization for crowd counting.

\section{Granular-Ball Graph Learning}

Graph-structured data are ubiquitous in complex systems such as social networks, bioinformatics, and knowledge graphs. Owing to their non-Euclidean nature and multiscale structural characteristics, they are often difficult to model effectively with traditional methods. Granular-ball computing provides a scalable multi-granular representation for graph data by representing local node sets as granular balls endowed with both geometric and topological meaning.

\subsection{Granular-Ball Graph Representation Model}

Granular-ball computing abstracts local node sets into granular balls with geometric and topological meaning. In supervised graph learning, a granular ball is typically represented as
\[
GB_i=\left(V_{GB_i},E_{GB_i},X_{GB_i},A_{GB_i},c_i\right),
\]
where \(V_{GB_i}\), \(E_{GB_i}\), \(X_{GB_i}\), and \(A_{GB_i}\) denote the node set, edge set, node-feature subset, and adjacency matrix within the granular ball, respectively, and \(c_i\) is the center node, which is usually chosen as the node with the highest degree within the ball. The granular-ball set usually satisfies full coverage over the original graph nodes, i.e., \(Cov(GB)=1\). On this basis, to balance representation quality and computational efficiency, granular-ball generation typically aims to use as few granular balls as possible while controlling the refinement level through quality constraints. In supervised settings, granular-ball quality is usually defined as the proportion of the dominant label within the ball. Based on the generalized granular-ball model (\ref{equ:GBC}), the following granular-ball graph representation model can be obtained:
\begin{equation}\label{graph}
	\begin{aligned}
		&\min_{c_i,V_{GB_i},E_{GB_i},m} \ -Comp(GB)+m \\
		& s.t. \quad quality(GB_i)\geq \phi(x), \ i=1,2,\ldots,m, \\
		& quality(GB_i)=\frac{\max(n_j)}{n}, \ j=1,2,\ldots,k,
	\end{aligned}
\end{equation}
where \(Comp(GB)\) characterizes the compactness of granular balls and is usually reflected by their internal connectivity strength; \(\phi(x)\) is the quality-threshold function, often specified as a fixed threshold \(T\); \(n_j\) denotes the number of nodes of class \(j\) within the granular ball; \(n\) is the total number of nodes in the granular ball; and \(k\) is the number of label classes.

In unsupervised graph learning, the formulation is similar to Eq.~(\ref{graph}), except that the quality function no longer depends on labels and is instead measured by the internal connection density of the granular ball~\cite{SGBGC2025}. By jointly controlling the number and quality of granular balls, granular-ball graph representation enables a multiscale partition of graph structures from global to local. In this way, it effectively preserves key topological information while reducing computational complexity, thus providing an efficient and interpretable representation for large-scale graph data analysis.

 \subsection{Generation and Optimization of Granular-Ball Graph Representation Models}
 
 The construction of granular-ball graph representation models depends not only on how granular balls are defined, but also on how they are generated and refined. A proper generation and optimization strategy is essential for achieving multigranular graph representation, reducing computational complexity, and preserving structural information. In supervised settings, granular-ball generation usually consists of two stages: coarse-grained partitioning and fine-grained partitioning~\cite{SGBGC2025}. The coarse-grained stage is used to quickly identify large structural units in the graph. Specifically, the node with the highest degree among the unassigned nodes is selected as the center, and breadth-first search (BFS) is then used to expand outward layer by layer to form the current granular ball. This process is repeated until approximately \(\sqrt{n}\) initial granular balls are obtained. In the fine-grained stage, the two highest-degree nodes within each parent granular ball are selected as splitting centers. Starting from these two nodes, BFS is again performed layer by layer, and the remaining nodes are assigned to the first child granular ball that reaches them. This procedure continues iteratively until the purity of all granular balls satisfies a predefined threshold, such as \(T=1\).
 
 In unsupervised graph learning, the generation of granular-ball graph representations generally follows the same coarse-to-fine two-stage framework~\cite{GBGC2025}. Unlike the supervised case, however, splitting is no longer driven by label purity, but by internal connectivity strength or structural compactness. By combining coarse-grained initialization with fine-grained adaptive refinement, the method progressively improves graph representation quality while maintaining efficiency, thereby providing a multigranular structural basis for unsupervised graph learning.

\subsection{Granular-Ball Graph Learning Models}

Although graph neural networks (GNNs) have achieved remarkable success in tasks such as node classification, graph classification, and graph regression, their computational and memory costs grow rapidly with graph size, making scalability a major bottleneck. To address this issue, Xia et al. proposed a granular-ball-based graph coarsening method~\cite{SGBGC2025}, which partitions the original graph into granular balls satisfying quality constraints and constructs a coarsened graph by treating these granular balls as super-nodes. In this way, the graph size is significantly reduced while key topological characteristics are largely preserved, thereby improving the training efficiency and scalability of GNNs. Further studies have shown that granular-ball graph coarsening can not only preserve the overall structure of the original graph in graph classification tasks, but may even improve classification performance in some cases~\cite{GBGC2025}. In addition, granular-ball ideas have been extended to multiview graph learning. For example, Wang et al. proposed GBCM-GCN, which integrates information from different views through granular-ball topology construction and collaborative matrix convolution, thereby obtaining higher-quality graph representations~\cite{wang2025capturing}.

Building on this line of work, granular-ball computing has further given rise to a series of graph learning models. For heterogeneous graph clustering, Zhao et al. proposed the adaptive granular-ball graph rewiring method AGGR~\cite{zhaoadaptive}, which improves overall graph homophily through multigranular granular-ball decomposition, adding edges within granular balls, and removing edges between them. For link prediction, Zhao et al. proposed the multigranular graph positional embedding method MGLP~\cite{Zhao2026Liu}, which constructs multigranular distance encodings through adaptive granular-ball partitioning, granular-ball trees, and hierarchical center graphs. For graph pooling, Xu et al. proposed the topology-preserving adaptive graph pooling method TPAGP~\cite{Zhao2026Gao}, which combines granular-ball partitioning with persistent homology to achieve topology-preserving multigranular pooling. For higher-order relation modeling, Zhao et al. proposed the multigranular hypergraph representation learning method MGHRL~\cite{Zhao2026Guan}, which generates multigranular hyperedges through granular balls and performs cross-granularity fusion by combining multichannel hypergraph networks with attention mechanisms. In addition, for self-supervised representation learning on heterogeneous graphs, Wang et al. proposed the multigranular graph contrastive learning framework GBGCL~\cite{Wang2026Xia}, which replaces random perturbation with granular-ball-based augmentation and constructs positive and negative sample pairs at different scales through coarse-to-fine granular-ball refinement, thereby improving the robustness and structure-awareness of heterogeneous graph representation learning.

\subsection{Other Granular-Ball Graph Learning Methods}

To address the low efficiency of full-batch training and the instability of sampling in large-scale GCN training, Cong et al. proposed the progressive granular-ball sampling fusion framework (PGBSF), which partitions the original graph into a series of subgraphs via granular-ball sampling and improves both training efficiency and classification performance through progressive training and parameter sharing~\cite{cong2024enhancing}. In addition, Chang et al. proposed the MIGC-CMamba framework, which combines multiscale time-series imaging, granular-ball clustering, and Mamba-based modeling for traffic flow prediction. By adaptively aggregating traffic-network nodes into multigranular granular balls, the method enables joint modeling of temporal features and spatial relations~\cite{chang2026migc}. These studies suggest that granular-ball computing can serve not only as a fundamental mechanism for graph compression and multigranular representation, but also as an effective component for graph explanation, graph sampling-based training, and spatiotemporal graph modeling, thereby continuously expanding the application scope of granular-ball graph learning.

\section{Interdisciplinary Research on Granular-Ball Computing}

Beyond its foundational learning models, granular-ball computing has been progressively extended to other theoretical and methodological domains, forming cross-disciplinary integrations with areas such as evolutionary computation, three-way decision, and quantum machine learning. These developments further broaden both the application scope and the methodological depth of granular-ball computing.

\subsection{Granular-Ball Evolutionary Computation}

Many optimization problems are characterized by high dimensionality, complexity, black-box objectives, strong nonlinearity, nonconvexity, multimodality, and dynamic variations, posing substantial challenges to traditional methods. Although evolutionary computation has developed a variety of classical frameworks, it still typically relies on point-wise individuals as the basic search units, and therefore often suffers from sparse sampling, limited structural information, and premature convergence. To address these issues, granular-ball evolutionary computation uses granular balls as regional representation units of the solution space, and performs hierarchical coarse-to-fine search through operations such as initialization, splitting, and selection. In this way, it preserves global search capability while improving robustness, adaptability, and computational efficiency in complex black-box optimization~\cite{xia2023granular}.

\subsection{Granular-Ball Blockchain}

While blockchain supports digital asset transactions and trustworthy collaboration, it also faces two major challenges: limited scalability and security vulnerabilities in smart contracts. Existing sharding and vulnerability detection methods often lack adaptability in dynamic, large-scale, and noisy environments, making it difficult to balance efficiency and robustness. As a computational paradigm consistent with the global-first, coarse-to-fine cognitive mechanism, granular-ball computing naturally supports both multigranular partitioning and dynamic aggregation of accounts, as well as hierarchical abstraction and structural modeling of code. Along this line, Wu et al.~\cite{wu2025gbshard} proposed an adaptive sharding method based on granular-ball partitioning, which performs coarse-to-fine granular-ball generation, refinement, merging, and migration over account transaction networks to achieve sharding construction and dynamic adjustment. For the problem of noisy labels in smart contract vulnerability detection, related studies have proposed robust detection methods based on granular-ball-enhanced contrastive learning, which mitigate the influence of noisy labels through stable semantic representation learning and granular-ball aggregation in semantic space. In addition, Wang et al.~\cite{wang2026robust} further proposed a vulnerability detection method based on granular-ball computing and weighted path-signature similarity. By combining key-path extraction, granular-ball aggregation of semantically similar identifiers, and weighted path-signature matching, the method enables more fine-grained and robust vulnerability identification.

\subsection{Granular-Ball Three-Way Decision}

Three-way decision~\cite{yao2010three}, rooted in decision-theoretic rough sets, partitions the universe into the positive, negative, and boundary regions, corresponding to acceptance, rejection, and deferred decision, respectively, and provides an effective mechanism for handling uncertainty. Granular-ball computing and three-way decision are naturally complementary. On the one hand, granular-ball representations enhance the ability of three-way decision models to handle continuous data adaptively; on the other hand, three-way decision offers an effective mechanism for managing uncertainty in granular-ball classifiers. Yang et al.~\cite{yang2023granular} were the first to use granular balls to represent equivalence classes of continuous data, and proposed granular-ball three-way decision and granular-ball sequential three-way decision models, thereby overcoming the difficulty of narrow-sense three-way decision in handling continuous data and the limitation of generalized three-way decision models that take individual samples rather than equivalence classes as inputs. Subsequently, for neighborhood-rough-set-based three-way decision models, where the input is a single neighborhood granule and thresholds must be preset, Yang et al. further developed a three-way classifier based on granular-ball neighborhood rough sets~\cite{yang20243wc}, which is more robust than conventional three-way classifiers and also outperforms existing granular-ball classifiers on uncertain data.

To address the problem that granular-ball \(K\)-nearest neighbor methods may suffer from imbalanced sample distributions within granular balls and a fixed \(K\) value, Yang et al.~\cite{yang2024adaptive} exploited the uncertainty-handling capability of three-way decision and proposed an adaptive-neighborhood three-way KNN classifier based on density granular balls, achieving substantial improvements in both efficiency and classification accuracy. In addition, Yang et al. proposed a sequential three-way decision method based on fuzzy granular-ball rough sets, namely S3WD-FGBRS~\cite{3WD-FGBRS2025}, and further developed a cost-sensitive three-way decision granular-ball generation method (CS3W-GBG)~\cite{yang2025cs3w} under the guidance of the justifiable granularity principle, where a granularity optimization mechanism is introduced to improve classification accuracy and robustness. Liu et al.~\cite{liu2024three} also incorporated three-way decision into granular-ball rough sets and proposed the three-way decision granular-ball rough set method (3WD-GBRS), which reduces the uncertainty risk of granular-ball classifiers by constructing appropriate multigranular spaces from the perspective of uncertainty.

Granular-ball three-way decision has also been extended to feature selection. Xia et al.~\cite{xia2024three} combined the efficiency of granular-ball computing with the interpretability of three-way approximations to construct a multigranular fuzzy entropy, and significantly improved feature-selection efficiency by partitioning both granular balls and the feature space into positive, negative, and boundary regions. For imbalanced data processing, Xie et al.~\cite{xie2024three} proposed a three-way hybrid granular-ball sampling method (TWHGBS), which effectively addresses imbalanced binary classification. In scenarios where data instances may carry uncertain labels, Yang et al.~\cite{Yang2025threeclassifier} developed a robust three-way classifier with shadowed granular balls. By incorporating information entropy, they first proposed an enhanced granular-ball generation method with adjustable granularity, and then, based on the principle of minimum uncertainty, used shadow mappings to divide granular balls into core, important, and unimportant regions for test samples. On top of these shadowed granular balls, a three-way classifier was established to distinguish certain classes from uncertain ones. Yang et al.~\cite{yang2026three} further proposed a robust granular-ball three-way classifier by combining coverage and specificity under the justifiable granularity principle. Moreover, the same principle has also led to the development of a three-way granular-ball support vector machine~\cite{GBODWiltold2025} and three-way granular-ball outlier detection methods~\cite{cheng2025gbmod}, both of which demonstrate significant advantages in classification accuracy, computational efficiency, and robustness.

\subsection{Quantum Granular-Ball Machine Learning}

In the context of the deep integration of big data and artificial intelligence, classical machine learning often faces high computational complexity and limited expressive power when handling high-dimensional, nonlinear, and large-scale data. Quantum computing offers a new computational paradigm, but in the NISQ era it is still constrained by the number of qubits and circuit depth. Granular-ball computing provides a natural mechanism for data reduction and multigranular modeling in quantum machine learning by compressing raw data into a small number of granular balls while preserving key geometric and statistical characteristics. Along this line, Yuan et al.~\cite{Yuan2026jia} proposed the PCA-guided granular-ball generation model P-GBG, which combines coarse sampling with PCA-based refinement to preserve geometric boundary information while reducing quantum resource consumption. Xia et al.~\cite{QGBKNN2025} further proposed the quantum nearest-neighbor method GB-QkNN, which jointly exploits granular-ball compression, HNSW indexing, and quantum parallelism to reduce the complexity of nearest-neighbor retrieval, and additionally designs circuits for granular-ball quantization and similarity computation. To further enhance the ability of granular-ball representations to capture complex global cluster structures and nonlinear data distributions, Yuan et al.~\cite{yuan2026xie} mapped granular-ball centers into a high-dimensional Hilbert space, replaced Euclidean distance with quantum similarity to measure relations among granular balls, and combined quantum kernel matrices with agglomerative hierarchical clustering to better characterize global cluster structures under complex nonconvex and entangled distributions. In this way, quantum granular-ball machine learning uses granular balls as a bridge between classical and quantum computation, compressing data scale while enhancing the expressive and computational power of quantum models.

\section{Discussion on the Main Scenarios Addressed by Granular-Ball Computing}

Granular-ball computing achieves multigranular representation by covering the sample space with granular balls, and its generation follows the global-first principle. Granular-ball quality evaluation is central to this framework. Although the specific measures vary across tasks, they essentially balance local representation error. Granular balls may take the standard spherical form or task-specific extensions, such as boundary-point or approximate rectangular forms, and can be generated either sequentially or through coarse-to-fine splitting. Overall, granular-ball computing offers efficiency, robustness, and interpretability. Its efficiency stems from elevating the computational unit from points to granular balls and adapting granularity to the data distribution; its robustness lies in its tolerance to noise and outliers; and its interpretability arises from the geometric and statistical characteristics of granular balls, such as coverage, purity, density, and radius. The discussion in this section is not restricted to specific tasks, but instead focuses on the data conditions, computational constraints, and objective requirements under which granular-ball computing is most likely to be effective, as well as the situations in which it should be used with caution or further improved. Accordingly, the following discussion is organized around its applicability, limitations and possible improvements, and future development opportunities.

\subsection{Main Application Scenarios and Applicability Boundaries of Granular-Ball Computing}

Granular-ball computing is primarily suited to large-sample scenarios. When sufficient samples are available, stable local group structures are more likely to emerge, making granular-ball splitting and quality evaluation more effective; at the same time, the efficiency gain from replacing point-wise computation with coarse-grained granular balls becomes more pronounced. In contrast, when data are sparse, local statistics become unstable, and excessive refinement may introduce unnecessary structural uncertainty. Typical application scenarios include:

(1) \textbf{Data organization and learning tasks with large samples, complex distributions, and nonconvex structures}: adaptive multigranular partitioning can quickly capture the global shape at a coarse granularity while further refining complex boundary regions, thereby preserving both global structure and local details. Typical examples include clustering, nearest-neighbor learning, density analysis, and anomaly detection.

(2) \textbf{Scenarios with significant noise and a need for robust learning}: granular balls describe local structure through group-level statistics, confining the influence of noise to local regions, and identify unstable areas through purity, density, and consistency, thereby enabling robust decision making. This is especially suitable for industrial data, sensor data, and weakly labeled data.

(3) \textbf{Resource-constrained engineering scenarios requiring reduced complexity while preserving structure}: by using fewer structural units in computation, granular balls reduce the cost of neighborhood search, distance computation, and density estimation. This is particularly beneficial in large-scale iterative tasks such as mean-shift, evolutionary computation, and clustering updates, and also supports real-time or lightweight deployment.

(4) \textbf{Scenarios emphasizing structural interpretation and visual analysis}: granular-ball covering provides a clear spatial decomposition and forms a coarse-to-fine explanatory chain, making it suitable for rule extraction, visual diagnosis, and structured evaluation.

\textbf{Applicability boundary}: when the sample size is small and approaches the dimensionality, local statistics become unstable and the reliability of granular-ball splitting and quality evaluation decreases. In such cases, the advantages of granular-ball computing are limited, and it often needs to be combined with dimensionality reduction, metric learning, or strong prior constraints. Granular-ball computing is therefore not a ``the finer the better'' paradigm; its effectiveness depends on the learnability of multigranular structures supported by sufficient samples.

\subsection{Current Limitations and Future Improvement Directions of Granular-Ball Computing}

Although the adaptive multigranular representation of granular-ball computing substantially improves efficiency, robustness, and interpretability, several issues still deserve further study.

(1) \textbf{The trade-off between the number of granular balls and generation efficiency}: when granular balls are split too finely, their number may increase rapidly, undermining efficiency and potentially weakening robustness. This issue also arises in granular-ball clustering and granular-ball graph representation. Future work should further optimize granular-ball generation under quality constraints, for example by introducing budget constraints, split-gain evaluation, joint split--merge optimization, and local selective refinement to suppress structural explosion.

(2) \textbf{Insufficient modeling of distance metrics and attribute weights}: most existing standard granular-ball representations are based on Euclidean distance and thus have limited adaptability to data with heterogeneous attributes or inconsistent attribute importance. Future research may incorporate attribute weighting, metric learning, and heterogeneous-feature fusion into granular-ball models, so that the center, radius, and even shape parameters of granular balls can be learned adaptively according to task requirements.

(3) \textbf{Adaptivity of minimum granularity and quality lower bounds}: some granular-ball methods still rely on a quality lower bound to control the minimum granularity, while fixed thresholds or simple adaptive strategies are often suboptimal. A more desirable solution would be to use data-dependent, stage-dependent, or region-dependent dynamic thresholds, so that different regions and different learning stages can employ more appropriate granularity control mechanisms.

(4) \textbf{Cross-domain applicability and domain-specific challenges}: although granular-ball computing has been extended to multiple areas, it still faces clear challenges in scenarios such as natural language processing. Text data exhibit strong sequential dependence and compositional structure, and geometric neighborhoods do not necessarily reflect semantic relations. In addition, limited sample sizes at the token level may weaken the stability of granular-ball quality evaluation. Future work may combine granular balls with sequence modeling, graph structures, or semantic structures, and construct granular-ball representations in sentence-embedding, phrase-embedding, or semantic-graph embedding spaces, together with sequential or semantic-consistency constraints, to improve cross-domain adaptability.

\subsection{Future Opportunities and Potential Breakthroughs}

Granular-ball computing has already established foundational theories and preliminary models in multiple areas of artificial intelligence, yet substantial room for development remains. On the one hand, more mature models can be extended under the current framework, such as granular-ball fuzzy sets, granular-ball neural networks, and granular-ball clustering, to further improve efficiency and robustness. On the other hand, for emerging scenarios involving noise robustness and large-scale computation, such as 3D point clouds, multi-label learning, and concept cognition, there is a clear need to develop reusable methodological frameworks.

Future progress may also come from method integration. For example, granular balls can be used as super-nodes in graphs to reduce computational cost, to characterize stable patterns and drift in time series, or to represent local consistency in multimodal alignment spaces. Such directions point toward a broader goal: integrating representation, measurement, and decision making within a unified multigranular structure, so that granular balls become reusable and extensible modules for structured computation.

\subsection{Acknowledgments}
\noindent This work was supported in part by the National Natural Science Foundation of China under Grant Nos. 62221005, 62450043, 62222601, 62176033, and Chongqing Natural Science Foundation Innovation and Development Joint Fund under Grant Nos. CSTB2025NSCQ-LZX0141.

\bibliographystyle{abbrv}                                                                             \bibliography{ref3}

\end{document}